\documentclass[shortAfour,sageh,times]{sagej}
\usepackage{amssymb}
\usepackage[]{amsfonts,amssymb}
\usepackage{algorithm,algorithmicx,algpseudocode}
\usepackage{undertilde}

\def\volumeyear{2019}

\newcommand{\LineSet}[3]{P^{\vec{#1},\vec{#2}}_{#3}}
\DeclareMathOperator{\atantwo}{atan2}
\DeclareMathOperator{\len}{len}
\DeclareMathOperator*{\argmin}{arg\,min}
\algnewcommand{\IIf}[1]{\State\algorithmicif\ #1\ \algorithmicthen}
\algnewcommand{\EndIIf}{\unskip\ \algorithmicend\ \algorithmicif}
\algnewcommand{\LineComment}[1]{\State \(\triangleright\) #1}

\begin{document}
\title{Robust Navigation of a Soft Growing Robot by Exploiting Contact with the Environment}
%
%
%

\author{Joseph D. Greer\affilnum{1},
        Laura H. Blumenschein\affilnum{1},
        Ron Alterovitz\affilnum{2},
        Elliot W. Hawkes\affilnum{3},\\
        and Allison M. Okamura\affilnum{1}}
        
\affiliation{\affilnum{1}Department of Mechanical Engineering, Stanford University\\
\affilnum{2}Department of Computer Science, University of North Carolina Chapel Hill\\
\affilnum{3}Department of Mechanical Engineering, University of California Santa Barbara}

\email{jdgreer@alumni.stanford.edu}

\begin{abstract}
Navigation and motion control of a robot to a destination are tasks that have historically been performed with the assumption that contact with the environment is harmful. This makes sense for rigid-bodied robots where obstacle collisions are fundamentally dangerous. However, because many soft robots have bodies that are low-inertia and compliant, obstacle contact is inherently safe. As a result, constraining paths of the robot to not interact with the environment is not necessary and may be limiting. In this paper, we mathematically formalize interactions of a soft growing robot with a planar environment in an empirical kinematic model. Using this interaction model, we develop a method to plan paths for the robot to a destination. Rather than avoiding contact with the environment, the planner exploits obstacle contact when beneficial for navigation. We find that a planner that takes into account and capitalizes on environmental contact produces paths that are more robust to uncertainty than a planner that avoids all obstacle contact.
\end{abstract}


\maketitle

\section{Introduction}
\begin{figure}[b]
\begin{center}
\includegraphics[width=\columnwidth]{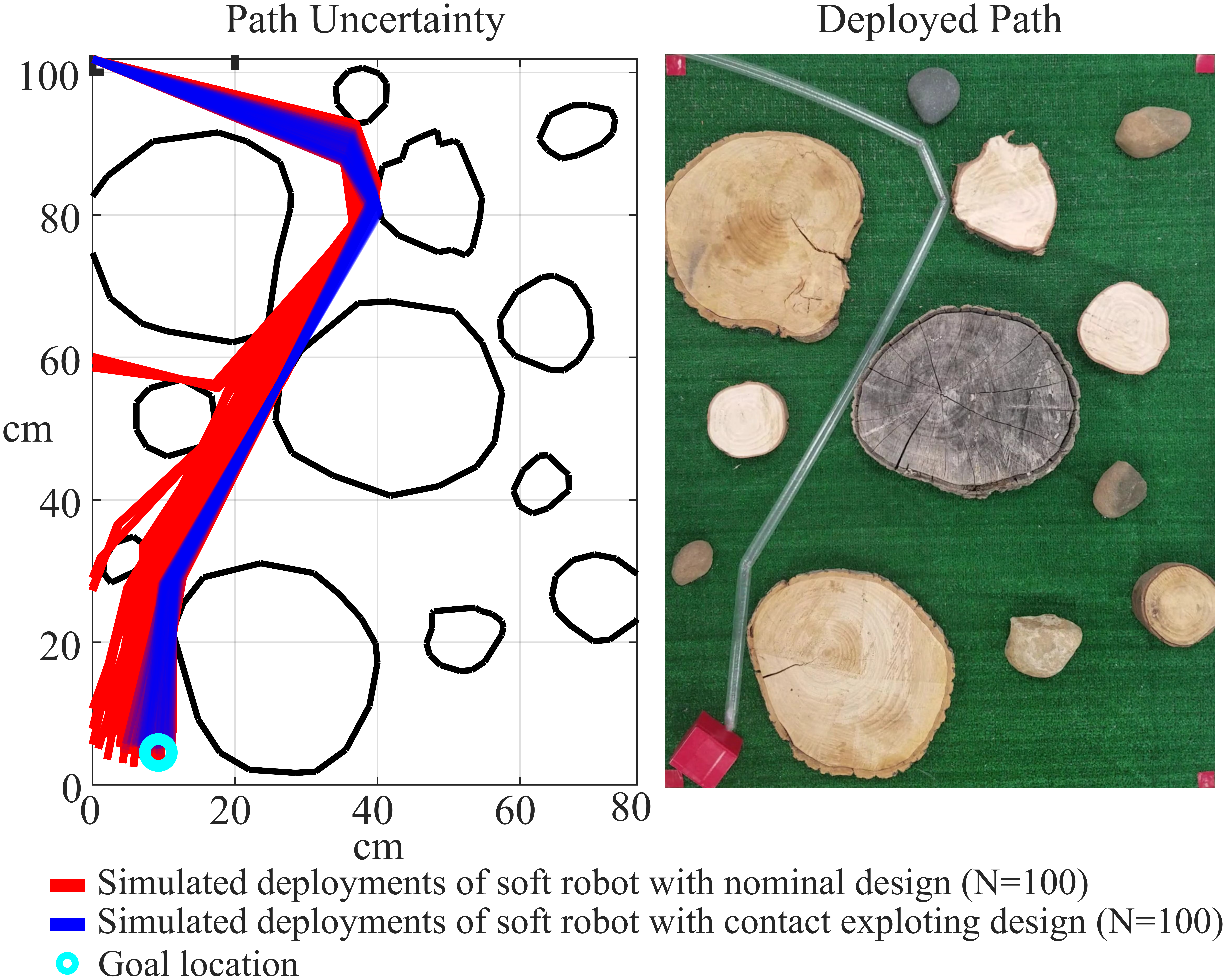}
\end{center}
\vspace{-6mm}
\caption[Obstacle Interaction Model Overview]{We present a heuristic model that enables planning of planar paths for a soft growing robot that exploit robot-obstacle interactions. Obstacles can be beneficial for navigation because they passively guide the robot and reduce uncertainty in its motion. Left: Simulated deployments of two robots with manufacturing uncertainty. One design (blue) is optimized to exploit obstacle contact to reach its goal location and the other design (red) is not. We illustrate 100 deployments of each design, showing that exploiting obstacle contact reduces uncertainty in the robot's path. Right: Physical deployment of the contact-exploiting design.}
\label{fig:Main} 
\vspace{-6mm}
\end{figure}

Soft robot bodies are made of materials that absorb and dissipate the energy of collisions, making environmental contact safe for many soft robots \citep{Rus2015}. In addition, the deformability of a soft robot's body causes it to naturally conform to the shape of an object it is in contact with. While these two properties have been demonstrated to benefit tasks that require object manipulation or gripping \citep{neppalli2007octarm,deimel2016novel,suzumori1992applying,katzschmann2015autonomous}, their benefits have not been explored for motion control and navigation of robots. Instead, these tasks have historically been performed with the assumption that contact with the environment is harmful. For applications such as search-and-rescue and inspection, which involve navigation of cluttered or constrained environments, environmental interaction may be unavoidable and even advantageous. In these scenarios, a system that allows for and exploits robot-obstacle interaction is desirable. 

A large body of research addresses modeling contact of a rigid body robot with the environment for the purposes of manipulation \citep{zheng1985mathematical,vukobratovic1999dynamics,killpack2016model}. For soft robotics, Coevoet et al. \citep{coevoet2017optimization} predict deformations of soft robots caused by environmental contact using numerical innovations to the finite element method that enable its use within a real-time control loop. In addition, Yip and Camarillo developed a model-less control strategy for tendon-based manipulators \citep{yip2014model}, which performs estimation of a tip Jacobian in an online fashion. Because the method does has not rely on an a priori model for the robot or environment, it can adapt to environmental contact. 

Other research examines using obstacles to the benefit of the robot. For example, hyper-redundant snake robots could actively use obstacles in their environment to propel the snake robot forward and thereby aid locomotion \citep{transeth2008snake,liljeback2009modelling,liljeback2012snake}. The authors developed both dynamic models for the interaction with obstacles and complementary control laws to utilize them in these papers. 
Another class of robot that uses environmental constraints to benefit mobility is pipe robots. It is only at pipe junctions that the robots have to make navigation decisions, otherwise they are directed along a path set by the shape of their environment \citep{roh2005differential}.

We consider interactions of a soft growing robot with obstacles, focusing in this paper on a soft robot that extends from its tip using pneumatically driven eversion \citep{HawkesScience}. In particular, we mathematically formalize obstacle interactions with the soft growing robot and use this formalization to plan paths for the robot to navigate to a destination. Similar to the recent works of \citet{pall2017} and \citet{sieverling2017} that consider the advantages of obstacle contact for motion planning, namely that it reduces uncertainty in the robot's motion, our planner generates paths that tolerate and even leverage obstacle collisions when helpful for navigating the soft growing robot to its destination. 

\begin{figure}[t]
\begin{center}
\includegraphics[width=\columnwidth]{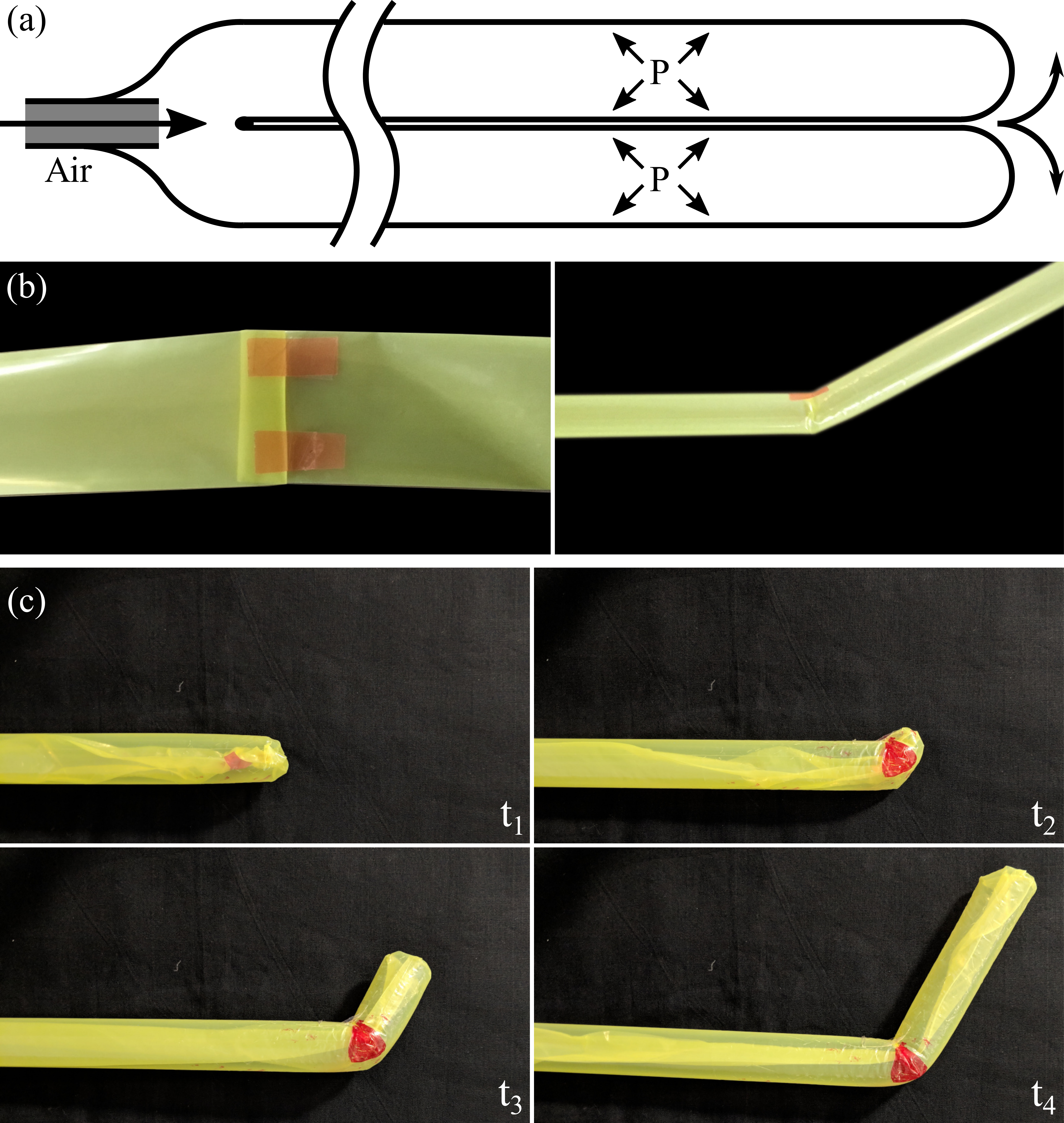}
\end{center}
\caption[Growing Robot Diagram]{Overview of the soft growing robot used in this work. (a) The robot extends from the tip using pneumatically driven tip-eversion \citep{HawkesScience}. (b) Turns are manufactured at discrete increments along the robot's body using tape to shorten one side of the robot relative to the other. (c) Sequence of four pictures shows the robot growing through a manufactured turn, which is marked in red. Due to the nature of tip extension, the turn's location does not move during growth.}
\label{fig:RobotOverview}
\vspace{-5mm}
\end{figure}

As in \citet{HawkesScience}, the robot considered turns via discrete pinches of its flexible plastic body along its length (\mbox{Fig. \ref{fig:RobotOverview}}). Because the soft growing robot's body does not slide with respect to its environment as it extends, the position of the turn does not move during the growth process once it has been everted. This paper extends \citet{GreerICRA2018}, which developed a model that predicted the motion of a soft growing robot moving through and interacting with a cluttered environment, by: (1) extending the model to handle designed turns in the robot's body, (2) developing a planner that exploits obstacle interactions, and (3) demonstrating the planner's performance in simulation and in experiments.

The remainder of this paper is divided into five sections. First, we develop a differential kinematic interaction model that describes infinitesimal motions of the robot when in contact with an obstacle. 
Second, we describe a method to plan paths to navigate to a destination using the obstacle interaction model developed in the previous two sections. Third, we present both simulation and physical experiments that validate the interaction model and planning method. The experiments demonstrate that the methods in this paper can be used to predict and plan trajectories of the soft growing robot through a planar environment with clutter. In the final section, we discuss implications and limitations of the methods and results presented in this paper.


\section{Planar Kinematic Model}
\label{sec:InteractionModel}
In this section, we develop a simple heuristic model that describes the differential kinematics of a soft growing robot that is moving through and potentially interacting with its environment. A soft growing robot consists of a pneumatic backbone that can extend in length as well as a turning mechanism that allows the soft growing robot to be steered from a straight-line trajectory to a destination. Several mechanisms have been proposed to steer a soft growing robot, including constant curvature bending of a robot's backbone caused by pneumatic artificial muscles that are attached along the pneumatic backbone's \mbox{length \citep{GreerICRA2017}} and asymmetric shortening of the robot's backbone at discrete intervals along its \mbox{length \citep{HawkesScience}}. In this paper, we consider the latter turning mechanism \mbox{(Fig.\ \ref{fig:RobotOverview})}, though the same ideas could be applied to other turning mechanisms and associated kinematic models that describe their motion in free space.

The soft growing robot in this paper belongs to the class of snake-like robots with flexible bodies known as continuum robots. Precise models of the kinematics and dynamics of continuum robots have been developed using continuum mechanics theory such as Cosserat rod \mbox{theory \citep{CalebRucker2014}} and the finite-element \mbox{method \citep{coevoet2017optimization}}. These methods are computationally expensive and rely on material parameters that may be difficult to estimate and change with time. A less exact, but simpler approximate modeling method that has been successfully used for certain continuum robots are lumped parameter models. These models characterize a continuum robot by specially chosen points along the robot's backbone. Examples of lumped parameter models of continuum robots include the unicycle model developed by \citet{Park2005} as well as the bicycle model developed by \citet{Webster2010}, both for steerable needles. We also use a lumped parameter kinematic model in this paper.

\begin{figure}[t]
\begin{center}
\includegraphics[width=\columnwidth]{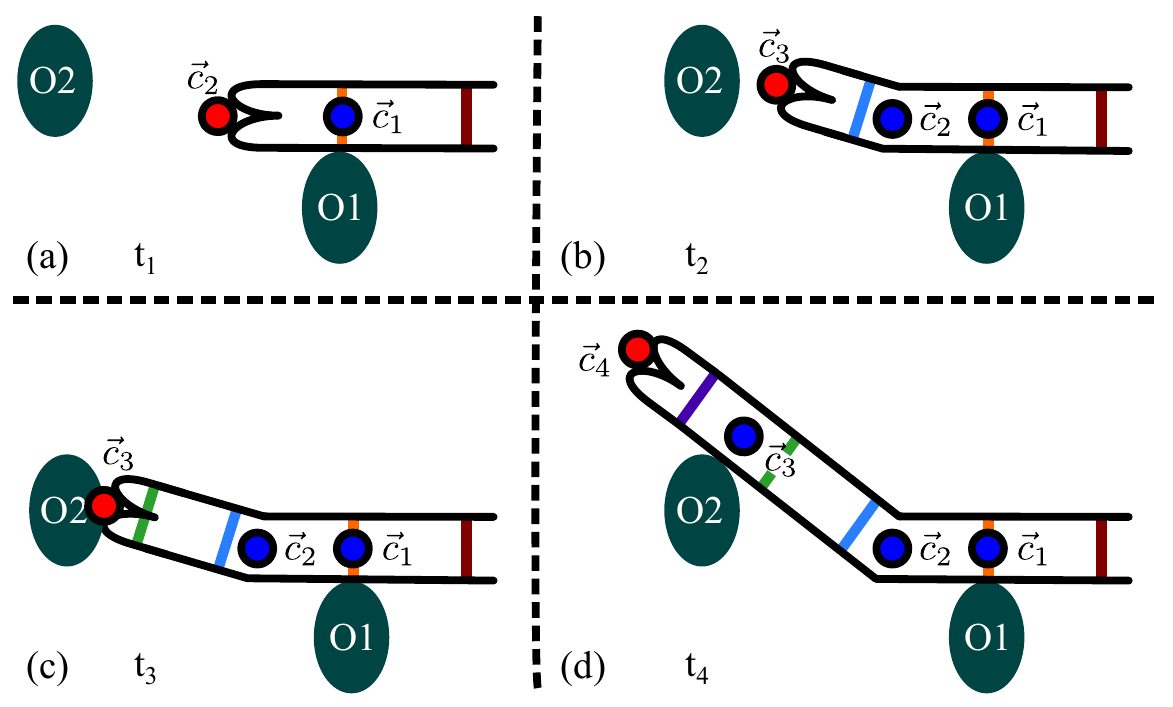}
\end{center}
\caption[Obstacle Interaction Lumped Parameter Model]{Model states of the lumped parameter model consist of pivot points $\vec{c}_1,\ldots,\vec{c}_n$. Obstacles in this figure are labeled O1 and O2. Pictures show the robot at four progressive time steps. From (a) to (b), a new pivot point is added as the robot turns at $\vec{c}_2$. Note that the position of the pivot does not change in subsequent time-steps because the body is extending, not translating (colored bands don't move). From (b) to (c) the robot's tip comes into contact with an obstacle. A new pivot point is not yet added. From (c) to (d) the robot's tip moves past the obstacle and a new pivot point is added.}
\label{fig:LumpedModel} 
\vspace{-5mm}
\end{figure}

\subsection{Model States}
\label{subsec:ModelStates}
Our lumped parameter model of the soft growing robot characterizes its state by specifically chosen points along the robot's backbone labeled $\vec{c_1},\ldots,\vec{c_n}$ \mbox{(Fig.\ \ref{fig:LumpedModel})}. Point $\vec{c}_n$, called the tip point, is defined as the position of the robot's tip. Points $\vec{c}_1,\ldots,\vec{c}_{n-1}$, called pivot points, are defined as the positions of the robot's backbone distinct from $\vec{c}_n$ that are either in contact with obstacles or positions where turns occur. If there is more than one contact point per obstacle, the model stores the most distal point of contact between each obstacle and the robot backbone. Note that the number of pivot points, $n$, varies as turns develop and the robot makes and breaks contact with its environment. In addition, a point of contact at a given obstacle is not added to the robot's state while the tip of the robot is in contact with the obstacle. The pivot points are ordered most proximal ($\vec{c}_1$) to most distal ($\vec{c}_n$), and the line segment from $\vec{c}_{n-1}$ to $\vec{c}_n$ represents the most distal segment of the soft growing robot.

\subsection{Joint Space Representation}
\label{subsec:CartJoint}
We elected to use a Cartesian space representation for the obstacle interaction model described in the Model States section, although in some cases it will be convenient to use a joint space representation instead. Therefore, we will briefly describe the joint space representation and how to convert to the joint space representation from the Cartesian space representation and vice versa.

The joint space representation of the robot consists of joint angles and corresponding segment lengths. Assume there is a robot with Cartesian space representation
\begin{equation}
(\vec{c}_1, \ldots, \vec{c}_n)
\end{equation}
and that, without loss of generality, $\vec{c}_1$ is coincident with the origin. The corresponding joint space representation will consist of $n-1$ segment lengths and $n-1$ joint angles:
\begin{equation}
(\theta_1, l_1,\ldots,\theta_{n-1}, l_{n-1}).
\end{equation}
The joint space representation can be computed from the Cartesian space representation as follows:
\begin{equation}
\label{eq:deltas}
\begin{array}{ll}
\vec{\delta}_i = \vec{c}_{i+1}-\vec{c}_{i} & \text{for } i = 1,\ldots,n-1\\
\theta_{i} = \atantwo(\delta_{iy},\delta_{ix}) & \text{for } i = 1,\ldots,n-1\\
l_i = ||\vec{\delta}_i|| & \text{for } i = 1,\ldots,n-1
\end{array}
\end{equation}
Similarly, the Cartesian space representation can be computed recursively from the joint space representation as
\begin{align}
\begin{array}{ll}
\vec{c}_{i+1} = l_{i} \hat{e}_{i+1} + \vec{c}_i & \text{for } i = 1,\ldots,n-1\\
\hat{e}_{i+1} = R_z(\theta_i) \hat{e}_i & \text{for } i = 1,\ldots,n-1\\
\vec{c}_1 = [0,0,0]^\top & \\
\vec{e}_1 = [1,0,0]^\top &
\end{array}
\end{align}
where $R_z(\theta)$ represents a z-axis rotation of $\theta$ radians. For the remainder of the paper, we will implicitly switch between representations when mathematically convenient, with the knowledge that they are different representations of the same information. When helpful, we will be explicit about transforming between representations using the following notation: 
\begin{align}
&(\vec{c}_1,\ldots,\vec{c}_n) = \text{CartesianSpace}(l_1,\theta_1,\ldots,l_{n-1},\theta_{n-1})\\
&(l_1,\theta_1,\ldots,l_{n-1},\theta_{n-1}) = \text{JointSpace}(\vec{c}_1,\ldots,\vec{c}_n)
\end{align}

\begin{figure}[t]
\begin{center}
\includegraphics[width=0.8\columnwidth]{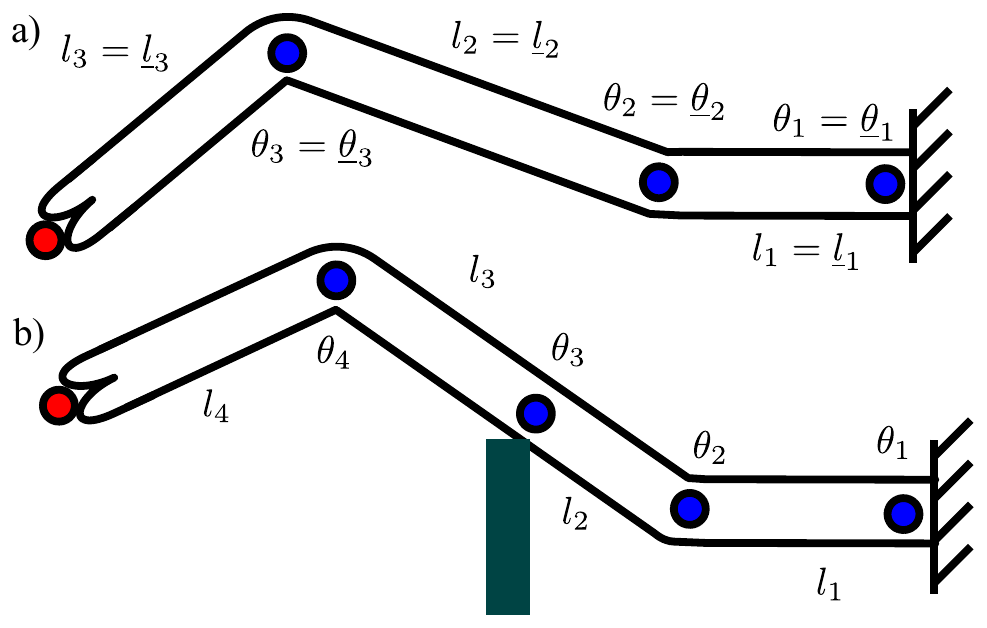}
\end{center}
\caption[Robot Design Diagram]{Robot design diagram. (a) A robot grown in the presence of no obstacles will end with a joint space state equal to its design parameters ($l_i = \underline{l}_i$ and $\theta_i = \underline{\theta}_i)$. (b) However, when the robot interacts with obstacles as it grows, the robot will end with a state that does not agree with its design parameters. In this case the robot has an extra pivot point and a larger deflection at joint 2 due to its interaction with the rectangular obstacle as it grows.}
\label{fig:RobotDesign} 
\end{figure}

\subsection{Robot Design}
\label{subsec:RobotDesign}
By placing turn points at selected locations along the robot's backbone at the time of manufacture, the robot can be designed to grow to destinations not reachable by a straight-line path.  Concretely, a robot design consists of a sequence of angular deflections, $\underline{\theta}_1,\ldots,\underline{\theta}_m$, and lengths between the turn points, $\underline{l}_1,\ldots,\underline{l}_m$ (\mbox{Fig.\ \ref{fig:RobotDesign}(a)}), where $\underline{\theta} \in [-\theta_M,\theta_M]$ represents the range of angular deflections that can be manufactured. For the remainder of the section, we will assume the robot is manufactured exactly as specified and manufacturing uncertainty will be considered in the next section.

The set of robot design parameters directly corresponds to a joint space state representation (Joint Space Representation section). Indeed, if the robot is deployed in the presence of no obstacles (and there are no errors in manufacturing), it will end with a state that has a joint space representation that matches the robot design parameters. Note that when the robot is deployed in free space, the position of a turn is fixed and independent of the length of the robot (i.e. the turn does not travel as the robot's length increases). When the robot is deployed in the presence of obstacles, it will interact with the environment, which will perturb the robot's state from its free space joint representation (\mbox{Fig.\ \ref{fig:RobotDesign}(b)}). The rest of this section is devoted to determining what this state will be, given a map that contains the locations of obstacles.

\subsection{Manufacturing Uncertainty}
\begin{figure}[t]
\begin{center}
\includegraphics[width=\columnwidth]{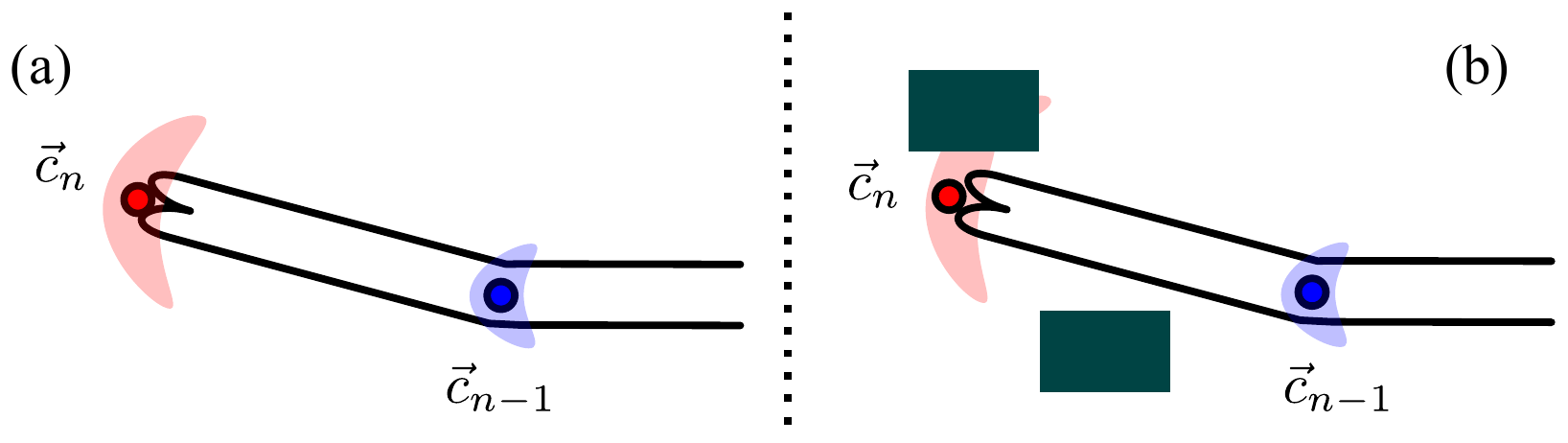}
\end{center}
\caption[Path Planning Uncertainty]{Errors in manufacturing and modeling lead to uncertainty in robot path. Representative regions of pivot point locations are depicted in light red and blue. In free space (a), the uncertainty compounds as expected, but in (b), the robot's path is bifurcated by the presence of an obstacle leading to two disjoint red regions of smaller area than in (a). Our planner exploits this reduction in area to increase the robustness of navigation. }
\label{fig:PathPlanningUncertainty}
\end{figure}

In practice, manufacturing a robot with nominal design $(\underline{l}_1,\underline{\theta}_1,\ldots,\underline{l}_m,\underline{\theta}_m)$ will not a yield a robot with the exact desired design parameters. Imprecision and errors in manufacturing will result in a robot with realized parameters $(\utilde{l}_1,\utilde{\theta}_1,\ldots,\utilde{l}_m,\utilde{\theta}_m)$ where $\utilde{l}_i,\utilde{\theta}_i$ are random variables distributed about the robot design parameters \mbox{(Fig.\ \ref{fig:PathPlanningUncertainty})} 
\begin{equation}
\label{eq:ProbQuantities}
\begin{aligned}
& \utilde{l}_i \sim \mathcal{U}(\underline{l}_i-\sigma_L, \underline{l}_i+\sigma_L) && \text{for } i = 1,\ldots,m \\ 
& \utilde{\theta}_i \sim \mathcal{U}(\underline{\theta}_i-\sigma_\theta, \underline{\theta}_i+\sigma_\theta) && \text{for } i = 1,\ldots,m 
\end{aligned}
\end{equation}
and $\sigma_L$ are $\sigma_\theta$ capture the tolerances of the manufacturing process. These parameters may be empirically estimated by measuring manufacturing errors for a population of robots. Alternatively, $\sigma_L,\sigma_\theta$ may be estimated by measuring the deviation between the deployed path and predicted path for a sample population of robots. This would provide a more holistic accounting of uncertainty, because it captures both manufacturing error and other sources of uncertainty, such as modeling error.

\subsection{Pivot Point Handedness}
As explained in the Model States section, pivot points, $\vec{c}_i$, correspond to obstacle contacts and manufactured turn points. Each pivot point also has a handedness associated with it. A pivot point associated with an obstacle contact is called a right (left) pivot point if rotating the robot to the right (left) about the pivot point moves the robot into the obstacle (using the right-hand rule). A pivot point associated with a manufactured turn is a right (left) pivot point if the angular deflection associated with the turn is negative (positive), $\underline{\theta} < 0 \text{  } (\underline{\theta} > 0)$. For example, in \mbox{Fig.\ \ref{fig:RobotDesign}(b)}, $\vec{c}_3$ and $\vec{c}_4$ are left-handed pivot points, and $\vec{c}_2$ is a right-handed pivot point. $\vec{c}_1$ corresponds to the robot's base, and therefore is both a right and left pivot point.

\begin{figure}
\begin{center}
\includegraphics[width=\columnwidth]{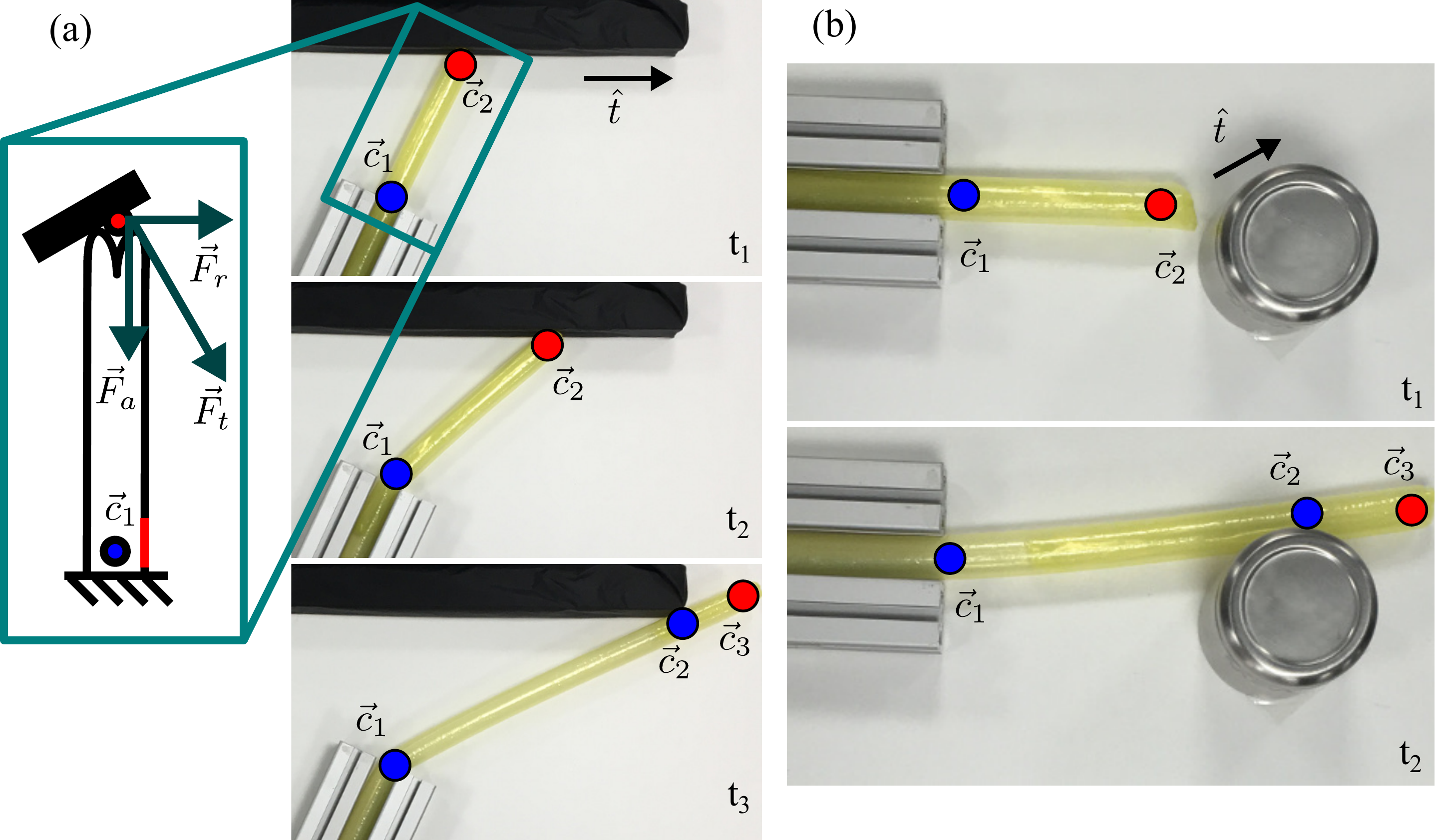}
\end{center}
\caption[Wall Obstacle Interaction]{Interaction of robot with an obstacle. (a) Robot shown interacting with a wall at three successive time-steps $t_1,t_2,t_3$. The robot comes into contact with the obstacle, after which the robot tip starts moving along direction $\hat{t}$, parallel to the obstacle surface, pivoting about point $\vec{c}_1$. After moving past the obstacle, the robot resumes free-growth kinematics with an updated pivot point, $\vec{c}_2$. The obstacle will exert a reaction force, $\vec{F}_r$, that has a transverse component. This will cause buckling about the pivot point, $\vec{c}_p$, at the red highlighted surface. In (b), an analagous sequence of interactions occur with a round obstacle. }
\label{fig:LargeObstacle}
\end{figure}

\subsection{Differential Kinematics of the Model}
\subsubsection{Free Growth}
\label{subsubsec:FreeGrowth}
In this section, we assume a known robot design $(\underline{l}_1,\underline{\theta}_1,\ldots,\underline{l}_m,\underline{\theta}_m)$. Given a robot with state $(\vec{c}_1,\ldots,\vec{c}_n)$, its length, which we denote $\len(\vec{c}_1,\ldots,\vec{c}_n)$, is the sum of the segment lengths $\len(\vec{c}_1,\ldots,\vec{c}_n) = \sum_{i=1}^{n-1} l_i$, where $l_i$ are the segment lengths of the joint space representation. We also let the $\underline{L}_i = \sum_{j=1}^i \underline{l}_j$ be the total backbone length at which designed turn $i$ will take effect (i.e. be everted by the robot).

Free growth occurs when the tip of the soft growing robot is not in contact with an obstacle. The tip of the robot will extend in the direction of the most distal segment of the backbone, $\hat{e}_{n-1}$, which is parallel to $\vec{c}_n-\vec{c}_{n-1}$. When $\len(\vec{c}_1,\ldots,\vec{c}_n) \neq \underline{L}_i$ for any $i$, we write the free growth differential kinematics simply as 
\begin{align}
&\dot{\vec{c}}_n = u  \hat{e}_{n-1}
\end{align}
where $u$ is the growth speed (rate of change of robot length), which we assume is controlled. Free growth is depicted at times $t_1, t_2,$ and $t_4$ of \mbox{Fig. \ref{fig:LumpedModel}}. When $\len(\vec{c}_1,\ldots,\vec{c}_n) = \underline{L}_i$ for any $i$, designed turn $i$ will be everted by the robot. This has the effect of instantaneously rotating the robot's tip heading by $\theta_i$ ($\theta_i$ is signed and the right-hand rule is used to determine direction of deviation). A new pivot point is added to the robot's state at the location the turn was everted
\begin{align}
&n = n+1 \\
&\vec{c}_n = \vec{c}_{n-1} \\
&\hat{e}_{n-1} = R_z(\theta_i) \hat{e}_{n-2}\\
&\dot{\vec{c}}_n = u \hat{e}_{n-1}
\end{align}
where $u$ and $\hat{e}_i$ have the same definitions as above. A designed turn is everted between time $t_1$ and $t_2$ in \mbox{Fig. \ref{fig:LumpedModel}}.


\subsubsection{Obstacle Contact}
\label{subsec:ObstacleContact}
In this section, we describe a model for movement of the soft growing robot when its tip is in contact with an obstacle. We assume that the robot will approach the obstacle in free growth as depicted in \mbox{Fig.\ \ref{fig:LargeObstacle}}(a). When it comes into contact with the obstacle, the soft growing robot will switch from free growing kinematics to obstacle contact kinematics. 

We treat the soft growing robot as an inflatable beam constrained at a previous pivot point, $\vec{c}_p$ for some $p$. How to determine $p$ will be explained below. The inflatable beam has a reaction force, $\vec{F}_r$, applied by the obstacle to the robot's tip. $\vec{F}_r$ acts normal to the obstacle surface, which is parallel to $\hat{t}$ and is shown in \mbox{Fig.\ \ref{fig:LargeObstacle}(a)}. The reaction force has components that are both transverse and axial with respect to the robot's backbone ($\vec{F}_t$ and $\vec{F}_a$, respectively), both of which could cause the inflated beam to buckle. $\vec{F}_t$ will cause a transverse beam buckling at the base \citep{masser1963deflections}, while $\vec{F}_a$ will cause an axial buckling near the center \citep{fichter1966theory}. The magnitude of the critical buckling force for each of these two modes depends on many parameters, such as the pressure in the inflatable beam, wall thickness of the robot's skin, length, and diameter. For pressures less than 15\,kPa, wall material of low density polyethylene with thickness on the order of 0.05\,mm, free length less than a meter, and diameter on the order of 20\,mm, transverse buckling at the base will occur in any case when the angle between the obstacle and the robot is greater than several degrees \citep{hammond2017pneumatic}.  

Once a bending moment that is larger than the tensioned wall and compressed air in the tube can resist is applied, the robot's backbone will buckle at pivot point $\vec{c}_p$. The net effect is that the tip of the robot will move tangentially to the obstacle's surface (parallel to $\hat{t}$), pivoting about point $\vec{c}_{p}$. The restoring moment at the point of buckling in the robot will ensure that its tip will remain in contact with the obstacle until it grows past the obstacle's edge. When this happens, the robot will switch back to free growth kinematics and a new contact point will be added to reflect the new point of contact between the robot and the obstacle it grew along. \mbox{Fig.\ \ref{fig:LargeObstacle}} shows the robot tip interacting with a wall and cylinder, respectively. The robot's tip flows around the obstacles, and once past, a new pivot point is added to reflect the new contact with the environment.

\begin{figure}[t]
\begin{center}
\includegraphics[width=\columnwidth]{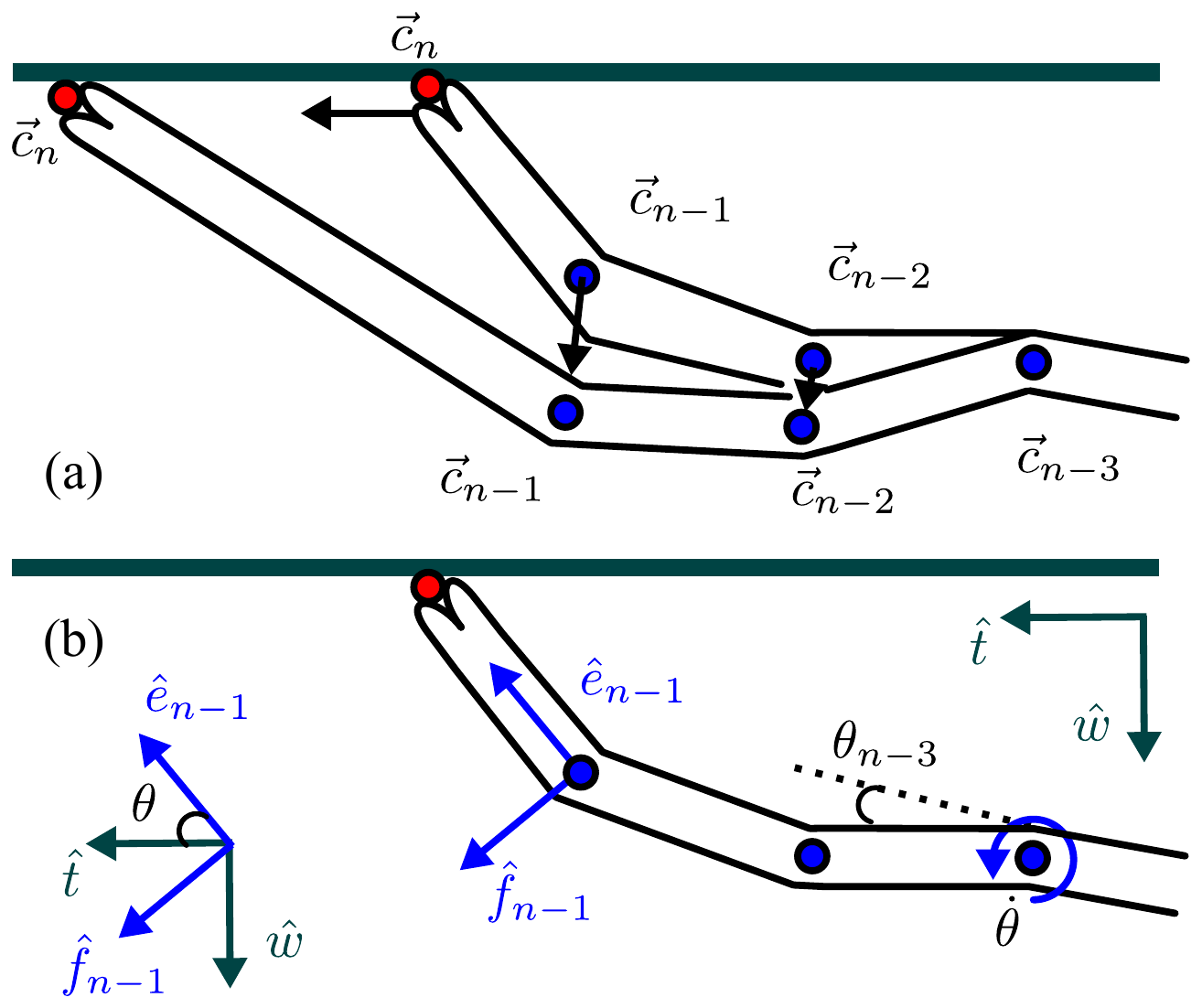}
\end{center}
\caption[Obstacle Interaction Diagram]{Soft Growing robot obstacle interaction kinematics. (a) Robot will pivot about most proximal unsupported pivot point that has the same handedness as the turn direction. In this case, the robot is being turned left by the obstacle and hence it will pivot about $\vec{c}_{n-3}$, the most proximal unsupported left pivot point. (b) Labels of relevant variables. $\hat{e}_{n-1} \times \hat{t} > 0$ so the robot is being turned left by the obstacle.}
\label{fig:ObstacleContactDiagram} 
\end{figure}

\paragraph{Obstacle pivot point $\vec{c}_p$:} As explained above, when the robot's tip is in contact with an obstacle, it will slide along the obstacle, pivoting about a previous pivot point in the robot's state ($\vec{c}_p$ for some $p$). The point about which it will pivot is the most proximal unsupported pivot point that has the same handedness as the direction the robot will be turned by interacting with the obstacle  (\mbox{Fig.\ \ref{fig:ObstacleContactDiagram}(a)}). This behavior results because previous pivot points have lower stiffness than unbuckled regions of the body, and the most proximal unsupported pivot point will have the highest moment of the pivot points. We define the direction the robot is turned by an obstacle as follows: As in \mbox{Eq.\ \ref{eq:deltas}}, define $\hat{e}_i$ to be the unit vector that is parallel to segment $i$ \mbox{(Fig.\ \ref{fig:ObstacleContactDiagram}(b))}. Assume $\hat{e}_{n-1} \cdot \hat{t}  > 0$. If not, negate $\hat{t}$. The robot will be pivoted to the left (right) if the z-component of $\hat{e}_{n-1} \times \hat{t}$ is positive (negative).

\begin{figure}[t]
\begin{center}
\includegraphics[width=1\columnwidth]{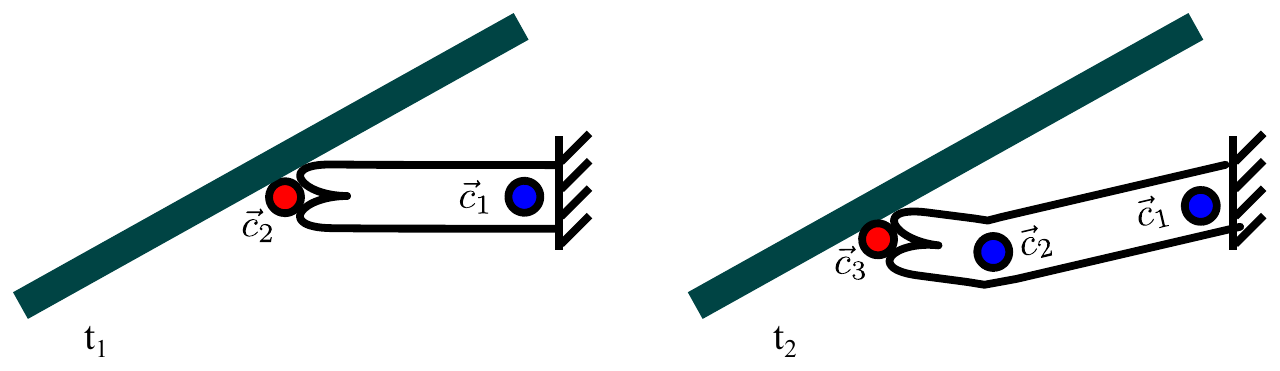}
\end{center}
\caption[Turning while in Contact with an Obstacle]{Illustration of the motion of the robot when a designed turn emerges while its tip is in contact with an obstacle. This is described mathematically by Eq. \ref{eq:DesignedTurnContact}.}
\label{fig:DesignedTurnContact}
\end{figure}

\paragraph{Obstacle Interaction Assumptions}To derive the obstacle interaction differential kinematics, we make the following assumptions: 

\begin{enumerate}
\item The robot's tip will follow the obstacle's tangent while the robot is in contact with the obstacle, i.e. $\dot{\vec{c}}_n$ is parallel to $\hat{t}$. 
\item Robot lengthening rate is a control input, i.e. $\dfrac{\mathrm{d}}{\mathrm{d} t}  \sum l_i = u$. 
\item New length is added to the last segment only, i.e. $\dot{l}_{n-1} = u$ and $\dot{l}_i = 0$ when $i \neq n-1$.
\item All joint angles except for the pivot point's joint angle remain constant, i.e. $\dot{\theta}_i = 0$ when $i \neq p$.
\end{enumerate}\vspace{0mm}
With these assumptions in place, we can write down the obstacle interaction differential kinematics in the joint-space representation in the general case:
\begin{align}
\begin{aligned}
&\dot{\theta}_i = 0, && i \neq p\\
&\dot{\theta}_p= \dot{\theta} && \\
&\dot{l}_i = 0, && i \neq n-1\\
&\dot{l}_{n-1} = u && 
\end{aligned}
\end{align}
where $\vec{c}_p$ is the obstacle pivot point and $\dot{\theta}$ is the unknown rotational velocity of the pivot point. When a designed turn is emerging, i.e.\ $\len(\vec{c}_1,\ldots,\vec{c}_n) = \underline{L}_j$ for some $j$, the obstacle interaction kinematics are as follows:
\begin{align}
\label{eq:DesignedTurnContact}
\begin{aligned}
&n = n+1\\
&\dot{\theta}_n = \delta \underline{\theta}_j\\
&\dot{\theta}_ i = 0, && i \neq p, i \neq n\\
&\dot{\theta}_p = \dot{\theta} && \\
&\dot{l}_i = 0, && i \neq n-1\\
&\dot{l}_{n-1} = u && 
\end{aligned}
\end{align}
which adds the designed discrete turn of $\underline{\theta}_j$ radians to the robot's state. As before, $\vec{c}_p$ is the obstacle pivot point (which may be changed as a result of the new designed turn) and $\dot{\theta}$ is the unknown rotational velocity of the pivot point. The motion described in this equation is depicted in \mbox{Fig. \ref{fig:DesignedTurnContact}}. The remainder of this section is devoted to determining $\dot{\theta}$.

To solve for $\dot{\theta}$, we first write $\dot{\vec{c}}_n$ in terms of $\dot{\theta}$ and use the first assumption, which says that the robot's tip velocity is parallel to the wall. We write the velocity of the robot's tip as follows
\begin{align}
\label{eq:TipVelFirst}\dfrac{\mathrm{d} \vec{c}_n}{\mathrm{d}t} &= \dfrac{^{\hat{e}_{n-1}} \mathrm{d} \vec{c}_n}{\mathrm{d} t} + \vec{\omega} \times \vec{r}^{\ \vec{c}_{n}/\vec{c}_{n-1}} + \dfrac{\mathrm{d} \vec{c}_{n-1}}{\mathrm{d} t}
\end{align}
where the first term is the robot's tip velocity expressed in the frame that is attached to point $\vec{c}_{n-1}$ and the second and third terms account for the linear and rotational velocity of the reference frame attached to $\vec{c}_{n-1}$. Using the assumption that all length is added to the last segment and all angular velocity is due to rotation about $\vec{c}_p$, we rewrite \mbox{Eq.\ \ref{eq:TipVelFirst}} as 
\begin{align}
\label{eq:TipVelSecond}
\dot{\vec{c}}_n &= u \hat{e}_{n-1}  + \dot{\theta} \hat{z} \times l_{n-1} \hat{e}_{n-1} + \dot{\theta} \hat{z} \times \vec{r}^{\ \vec{c}_{n-1} / \vec{c}_p}\\
\label{eq:TipVelLast}
&= u \hat{e}_{n-1} + \dot{\theta} \hat{z} \times \left(l_{n-1} \hat{e}_{n-1} + \vec{r}^{\ \vec{c}_{n-1} / \vec{c}_p}\right)
\end{align}
Finally, we enforce that the robot's tip velocity is parallel to the wall tangent 
\begin{equation} 
\label{eq:TipVel}
u \hat{e}_{n-1} + \dot{\theta} \hat{z} \times (l_{n-1} \hat{e}_{n-1} + \vec{r}^{\ \vec{c}_{n-1} / \vec{c}_p}) = v \hat{t}
\end{equation}
where $v$ is the unknown magnitude of the tip velocity. Since $\hat{e}_{n-1}$ and $\vec{r}^{\ \vec{c}_{n-1} / \vec{c}_p}$ have no $\hat{z}$ component, this equation reduces to two scalar equations that allow us to solve for the two unknowns, $\dot{\theta}$ and $v$.

\section{Planning Paths that Exploit Obstacles}
\label{sec:InteractionPlanning}

The Planar Kinematic Model section established a model that predicts the motion of the soft growing robot moving through and interacting with an environment. This section builds on this model to plan paths for the soft growing robot through an environment with obstacles. For robots moving through cluttered environments, it is inevitable that the robot will interact with obstacles. Rather than strictly avoiding environmental contact as is a standard paradigm for traditional path planning \citep{lavalle2006planning, choset2005principles}, the planner may decide to allow obstacle contact if it helps the soft growing robot reach its destination. For example, interactions with obstacles can consolidate many possible paths down to a single path, thereby reducing uncertainty \citep{pall2017, sieverling2017}. These interactions can direct the robot to locations not on a straight line path from its starting point, also reducing the need for designed turns that increase robot complexity. 

The planner can be logically divided into two parts. First, it determines a sequence of waypoints (defined in the Map Waypoints section) through which the soft growing robot will pass. Second, the planner generates a robot design that realizes the waypoint sequence with maximum expectation given the uncertainties described in \mbox{Eq.\ \ref{eq:ProbQuantities}}.

\subsection{Problem Definition}
\label{subsec:PPPrelims}

We assume the following information is provided:
\begin{enumerate}
\item A discretization of $\mathbb{R}^2$, $Z$.
\item A planar map, $M\subset Z$ that contains the locations of all obstacles in the map. $\vec{p} \in M$ if and only if $\vec{p}$ is part of an obstacle. 
\item An obstacle boundary vertex set, which consists of vertices of the obstacles in the map (Fig. \ref{fig:WaypointSequence}).
\item Starting point of the robot: $\vec{x}_s \in Z$
\item Goal point to navigate the tip of the robot to, \mbox{$\vec{x}_d \in Z$}, and radius, $d$, such that the task is considered successful if the robot's tip is within a distance $d$ of the goal point.
\end{enumerate}

The objective of this section is to produce a nominal robot design that causes the robot's tip to reach the destination with maximum expectation given the sources of uncertainty described in the Manufacturing Uncertainty section. Concretely, our objective is to find a nominal robot design, $(\underline{l}_1,\underline{\theta}_1\ldots,\underline{l}_{n-1},\underline{\theta}_{n-1})$, such that a robot manufactured according to that design, $(\utilde{l}_1,\utilde{\theta}_1,\ldots,\utilde{l}_{n-1},\utilde{\theta}_{n-1})$ has a high expectation of reaching the destination:
\begin{equation}
E\left(||\vec{x}_d - \utilde{\vec{c}_n}|| < d \right)
\end{equation}
where $(\utilde{\vec{c}_1},\ldots,\utilde{\vec{c}_n})$ is the deployed robot state computed using the obstacle interaction model.

\subsection{Map Waypoints}
\label{subsec:MapWaypoints}
\begin{figure}[t]
\begin{center}
\includegraphics[width=0.8\columnwidth]{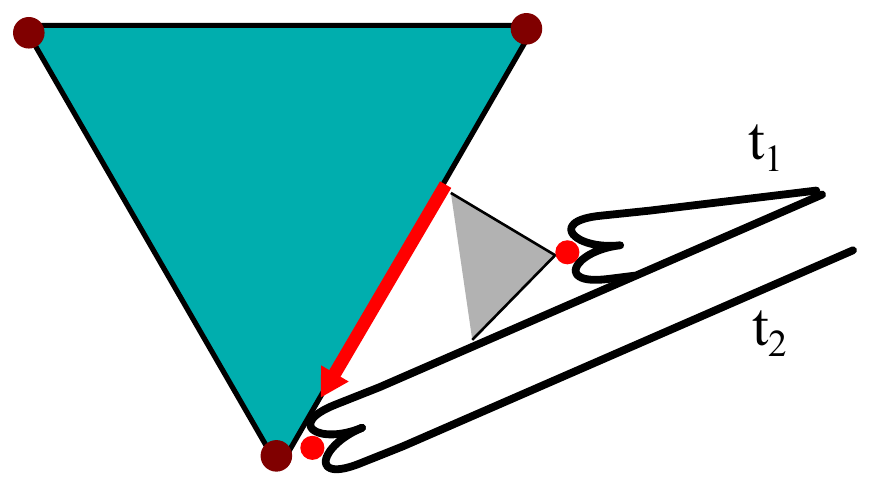}
\end{center}
\caption[Condensation Point Illustration]{Condensation point illustration. Due to the nature of obstacle interaction, obstacles condense many incoming robot paths through one of its vertices. In this example, all robots that approach the obstacle with a tangent that lay within the gray angle range are directed by the obstacle to its bottom vertex.}
\label{fig:CondensationPoints} 
\vspace{-5mm}
\end{figure}
As depicted in \mbox{Fig.\ \ref{fig:LargeObstacle}}, the obstacle interaction model predicts a simple behavior from the robot when it comes into contact with a polygonal obstacle: the tip will slide along the obstacle's contour until reaching a corner of the obstacle. This means that all incoming paths to an obstacle will be passively guided through one of the obstacle's vertices \mbox{(Fig.\ \ref{fig:CondensationPoints})}. Therefore, obstacle vertices are natural decision points for the planner because they are easy locations to reach. Note that this behavior closely links the path of the soft growing robot in the map with the map's visibility graph \citep{lozano1979algorithm}. In particular, when there are no designed turns, the body of a deployed robot will have the same shape as a path in the map's visibility graph.

We take advantage of this feature of the obstacle interaction model by using an approach inspired by a sampling-based roadmap \citep{kavraki1996probabilistic} in which our planner will generate robot designs that turn at only a subset of $x$-$y$ locations, which we call the waypoint set and denote by $W \subset Z$. Given a map, $M$, $W$ consists of the vertices of the obstacles in $M$ and is augmented with selected or random interior points, as well as the start and destination points (\mbox{Fig.\ \ref{fig:WaypointSequence}}). 

\subsection{Waypoint Sequence Generation}
\label{subsec:WaypointGeneration}
\begin{figure}[t]
\begin{center}
\includegraphics[width=0.8\columnwidth]{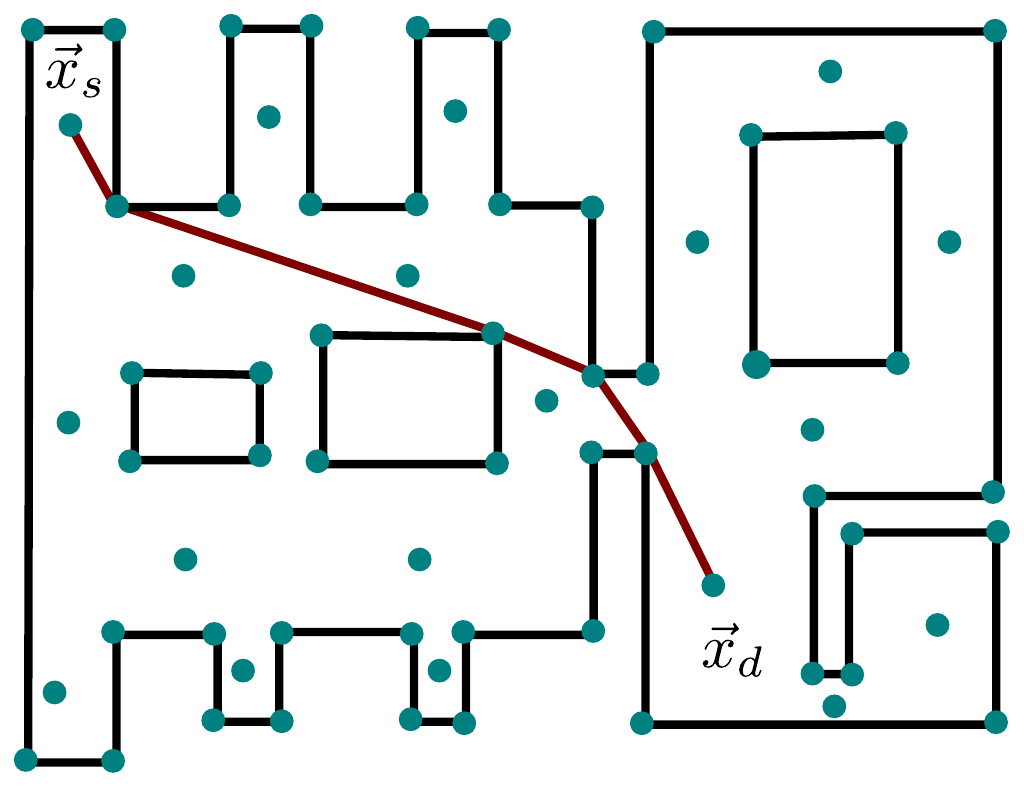}
\end{center}
\caption[Waypoint Sequence Generation]{Map with waypoint sequence illustration. A waypoint set consists of positions marked with solid teal circles. The waypoint sequence is shown by the maroon line, which connects the minimum turn waypoint sequence from $\vec{x}_s$ to $\vec{x}_d$ in series. This line does \textit{not} show the optimal robot design, which is calculated in the Robot Design Generation section.}
\label{fig:WaypointSequence} 
\end{figure}
The first part of the planner generates a sequence of points $( \vec{x}_1, \ldots, \vec{x}_N ) \in W^N$ through which the tip of the robot will pass in between the start and destination points (\mbox{Fig. \ref{fig:WaypointSequence}}). Because uncertainty is added with the addition of each designed turn, the objective of the planner is to choose a sequence of waypoints that allows the robot to reach its destination with the fewest number of designed turn points, relying as much as possible on passive redirection by the environment.

\begin{figure}
\begin{center}
\includegraphics[width=\columnwidth]{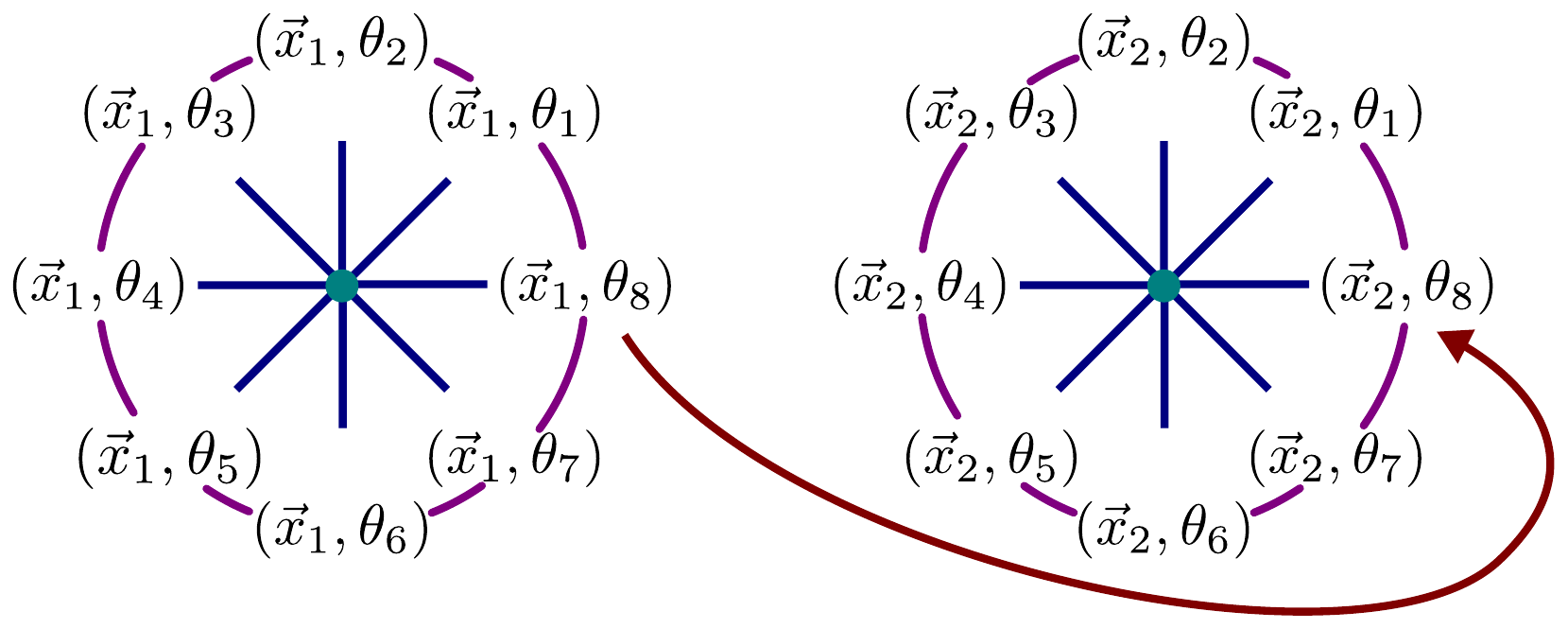}
\end{center}
\caption[Graph Structure]{ Structure of graph for waypoint sequence generation problem. Nodes of the graph represent departing from a waypoint, $\vec{x} \in W$ with a tip angle $\theta \in [0,2\pi)$.  Edges between two nodes exist if: (1) The two nodes correspond to the same waypoint and have tip angles that are within $\theta_M$ radians of one another. These edges represent adding a turn to the design (shown as purple lines). (2) A robot that departs from $\vec{x}_1$ with tip angle $\theta_1$ arrives at $\vec{x}_2$ with tip angle $\theta_2$ (shown as a maroon line).}
\label{fig:GraphStructure}
\vspace{-5mm}
\end{figure}

We define a directed graph whose structure is depicted in \mbox{Fig.\ \ref{fig:GraphStructure}}. Nodes of the graph consist of landmark points and tip angles, \mbox{$(\vec{x}, \theta) \in W \times T$}, where $T$ is a discretization of $[0, 2 \pi)$. Consider a node $(\vec{x}_1, \theta_1) \in W \times T$. If a path calculated using the obstacle interaction model with initial tip position $\vec{x}_1$ and tip angle $\theta_1$ with no designed turns goes through $\vec{x}_2$ with tip angle $\theta_2$, then an edge of weight 0 is added from $(\vec{x}_1,\theta_1)$ to $(\vec{x}_2,\theta_2)$. In addition, bidirectional edges of weight 1 that represent designed turns are added between all graph nodes $(\vec{x}_0, \theta_1)$ and $(\vec{x}_0, \theta_2)$ that represent the same landmark and satisfy the property that $\theta_1$ and $\theta_2$ are within $\theta_M$ radians of one another. (Recall that $\theta_M$ is the maximum angular deflection manufacturable.) Finally, a special start node, $\mathbf{s}$, and end node, $\mathbf{e}$, are added to the graph. Zero-weight edges are added from $\mathbf{s}$ to $(\vec{x}_s,\theta)$ for all $\theta \in T$ as well as zero-weight edges $(\vec{x}_d, \theta)$ to $\mathbf{e}$ for all $\theta \in T$.

The problem of finding a path with minimum number of designed turns is solved by finding a shortest path from the start node, $\mathbf{s}$, to the end node, $\mathbf{e}$. In our implementation, we used Dijkstra's algorithm to solve for the shortest path. Let \mbox{$( \mathbf{s},(\vec{x}_s,\theta_0),(\vec{x}_1,\theta_1),\ldots,(\vec{x}_N,\theta_N),(\vec{x}_d,\theta_{N+1}),\mathbf{e} )$} be the shortest path from node $\mathbf{s}$ to node $\mathbf{e}$. Note that such a path in the graph exists if and only if the end point is reachable from the start point by the robot through the map waypoints. Because the only edges that have non-zero weight represent turns, the minimum weight path from $\mathbf{s}$ to $\mathbf{e}$ represents a sequence of waypoints through which the robot will pass with the fewest number of designed turns. The waypoint sequence is given by ignoring the angles of the nodes in the shortest path: $(\vec{x}_s,\vec{x}_1,\ldots,\vec{x}_N,\vec{x}_d)$.

\subsection{Robot Design Generation}
\label{subsec:RobotDesignGeneration}

\begin{algorithm}
\caption{Optimal Robot Design} 
\label{alg:OptimalDesign}
\raggedright\hspace*{\algorithmicindent} \textbf{Input} Incremental Robot Design to $(i-1)$th waypoint, $i$th waypoint, Map\\
\raggedright\hspace*{\algorithmicindent} \textbf{Output} Incremental Robot Design to $i$th waypoint\\
\begin{algorithmic}[1]
\Procedure{OptimalDesign}{$[\underline{l}_1,\underline{\theta}_1,\ldots,\underline{l}_j,\underline{\theta}_j], \vec{x}_i,  M$}
\State $optimalSuccess \gets 0,\  \underline{\theta}_{j+1} \gets 0,\ \underline{l}_{j+1} \gets 0$
\ForAll{$\theta \in T$}
	\State Generate samples of $(\utilde{l}_1,\ldots,\utilde{\theta}_j)$
	\State Throw out samples that don't result in path that ends near $\vec{x}_{i-1}$
	\If{$|\theta| > 0$}
		\State Generate samples of $\utilde{\theta}$
	\EndIf
	\State $ESuccess \gets $ \# samples that end near $\vec{x}_i$/\# of samples
	\If{$ESuccess > optimalSuccess$}
	 	\State $optimalSuccess \gets ESuccess,\  \underline{\theta}_{j+1} \gets \theta$
	 	\State $\underline{l}_{j+1} \gets $ sample average of $\len(\utilde{\vec{c}}_1,\ldots,\utilde{\vec{c}}_k) - \sum_{\alpha=1}^j \underline{l}_\alpha$
	\EndIf
\EndFor
\EndProcedure
\end{algorithmic}
\vspace{-5mm}
\end{algorithm}
The second part of the planner generates a nominal robot design, $(\underline{l}_1,\underline{\theta}_1,\ldots,\underline{l}_m,\underline{\theta}_m)$, that results in a robot that reaches the destination with maximum expectation through the sequence of waypoints, $(\vec{x}_s,\vec{x}_1,\ldots,\vec{x}_N,\vec{x}_d)$, generated in the first part of the planner. For the sake of computational efficiency, we use a greedy approach to generate the robot design, in which we incrementally generate a design that maximizes the expectation of reaching the $i$th waypoint, given the tip of the robot is at the $(i-1)$th waypoint with robot design $(\underline{l}_1,\underline{\theta}_1,\ldots,\underline{l}_{j},\underline{\theta}_{j})$ for some $j <= i-1$. ($j = i-1$ when the optimal design has the robot turn at every waypoint). We do this by choosing $\underline{l}_{j+1},\underline{\theta}_{j+1}$ as follows
\begin{multline}
\label{eq:GreedyOptimization}
\underline{l}_{j+1},\underline{\theta}_{j+1} = \\ \underset{l,\theta}{\text{argmax }} \text{E}\left(||\vec{x}_i - \utilde{\vec{c}}_i|| < d \  \middle| \  ||\vec{x}_{i-1}-\utilde{\vec{c}}_{i-1}|| < d \right)
\end{multline}
where $\utilde{\vec{c}}_{i-1}$ and $\utilde{\vec{c}}_i$ are realized pivot point random variables derived from the nominal design $(\underline{l}_1,\underline{\theta}_1,\ldots,\underline{l}_{j+1},\underline{\theta}_{j+1})$. The decision not to turn at $\vec{x}_i$ is represented by the solution $\underline{\theta}_{j+1} = 0$ in \mbox{Eq.\ \ref{eq:GreedyOptimization}}.

\mbox{Alg.\ \ref{alg:OptimalDesign}} describes our method for generating the optimal design through waypoints $\vec{x}_1,\ldots,\vec{x}_{i}$. We use a collection of particles to represent probability distributions in this work, where each particle is a sample from the random variable being represented \citep{thrun2000monte,melchior2007particle}. This allows us to approximate the probabilistic quantities in \mbox{Eq.\ \ref{eq:GreedyOptimization}} using sample-based approximations. Let $(\underline{l}_1,\underline{\theta}_1,\ldots,\underline{l}_{j},\underline{\theta}_{j})$ be the nominal design corresponding to waypoint $i-1$. We generate particles by sampling $\utilde{l}_k,\utilde{\theta}_k$, for $k=1,\ldots,j < i$, according to the distributions given in \mbox{Eq.\ \ref{eq:ProbQuantities}}. Then the obstacle interaction model is used on each sampled robot design to compute the corresponding samples from the distribution of pivot point states, $(\utilde{\vec{c}}_1,\ldots,\utilde{\vec{c}}_K)$, where $K$ is the number of pivot points of the deployed robot. Since we are interested in the conditional expectation of states that reach the $(i-1)$th waypoint in \mbox{Eq.\ \ref{eq:GreedyOptimization}}, samples that are not within $d$ units of $\vec{x}_{i-1}$ are thrown out (shown on line 5 of \mbox{Alg.\ \ref{alg:OptimalDesign}}). $d$ is a pre-defined distance threshold that is used to demarcate successfully reaching a location.

To find the optimal next segment length, $\underline{l}_{j+1}$, and turn angle, $\underline{\theta}_{j+1}$, to reach $\vec{x}_i$, we evaluate the expectation in \mbox{Eq.\ \ref{eq:GreedyOptimization}} for every $\theta \in T$, where $T$ is a discretization of $[-\pi_M, \pi_M]$. In particular, for each $\theta \neq 0$, we sample $\utilde{\theta} \sim \mathcal{U}(\theta-\sigma_\theta,\theta+\sigma_\theta)$ and use the obstacle interaction model to determine if the resulting path goes near the $i$th waypoint. This allows us to evaluate a sample-based approximation of the expectation in \mbox{Eq.\ \ref{eq:GreedyOptimization}} and therefore the optimal $\underline{\theta}_{j+1}$. $\underline{l}_{j+1}$ is computed as the sample average of the difference between the robot length at waypoint $\vec{x}_i$ and the length of the robot at waypoint $\vec{x}_{i-1}$ for the optimal turn amount, $\underline{\theta}_{j+1}$. This is shown on line 12 of \mbox{Alg.\ \ref{alg:OptimalDesign}}. When $\theta = 0$, we do not sample $\utilde{\theta}$ since additional uncertainty is added only when new turns are made. This has the effect of favoring designs with fewer turns.

\section{Experimental Results}
\label{sec:ExperimentalResults}
In this section, we describe experiments that were performed to test the obstacle interaction model and path planning algorithm presented in the previous section. 
We start with experiments that test obstacle interactions with basic shapes such as walls and circles. Next, we present tests of the obstacle interaction model in a more complex scenario that chains multiple obstacle interactions together. These results review and extend the work presented in \citep{GreerICRA2018}. We then present both numerical and physical experiments that test the planning method presented in this work.

\subsection{Growth Along a Wall}
\label{subsec:WallGrowth}

\begin{figure}[t]
\begin{center}
\includegraphics[width=1\columnwidth]{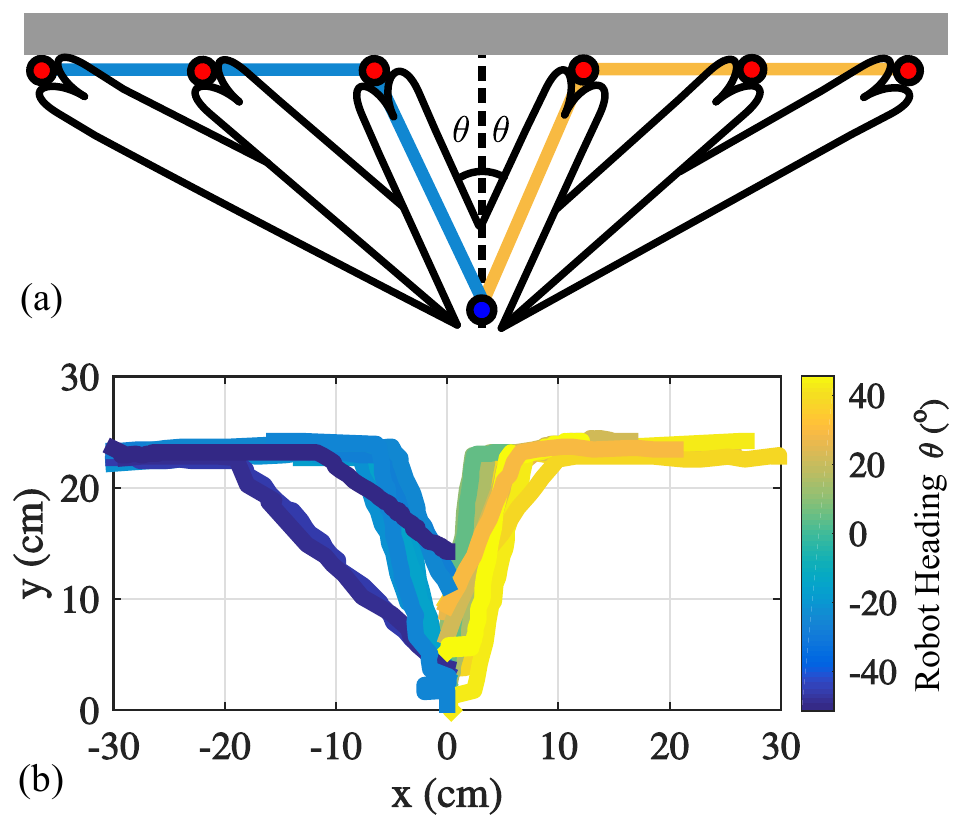}
\end{center}
\caption[Wall Tip Trajectories]{Experimental trajectories of robot tip position when the robot comes into contact with a wall. (a) Schematic showing two example trajectories and relevant parameters. (b) Experimental trajectories from 20 trials of the robot. As predicted by the model, the tip follows the wall trajectory, to the right if $\theta$ is positive, to the left if negative. Adapted from \citep{GreerICRA2018} \textcopyright\ IEEE 2018}
\label{fig:TrajectoriesWall} 
\end{figure}

As explained in the Map Waypoints section, the obstacle interaction model predicts that when the tip of the robot is in contact with an obstacle, it will slide along the object's contour in the direction most tangent to its distal segment while rotating about the obstacle pivot point. We performed two experiments to test this model. In the first experiment, we repeatedly grew the robot toward a wall 
from different approach angles. An overhead camera was used to capture the trials. Using color-based image segmentation, we extracted the position of the robot's tip over the course of each trial growth, forming a tip trajectory. \mbox{Fig.\ \ref{fig:TrajectoriesWall}(a)} illustrates two example starting angles, with the paths the obstacle interaction model predicted and their corresponding tip trajectories in colored lines. \mbox{Fig.\ \ref{fig:TrajectoriesWall}(b)} shows the results of the experiment, with 20 trial growths. Trajectories were colored by approach angle ($0^\circ$ corresponding to perpendicular to the wall). As predicted, the robot tip slid along the wall in the direction most tangent to its approach angle. 

\begin{figure}[t]
\begin{center}
\includegraphics[width=0.9\columnwidth]{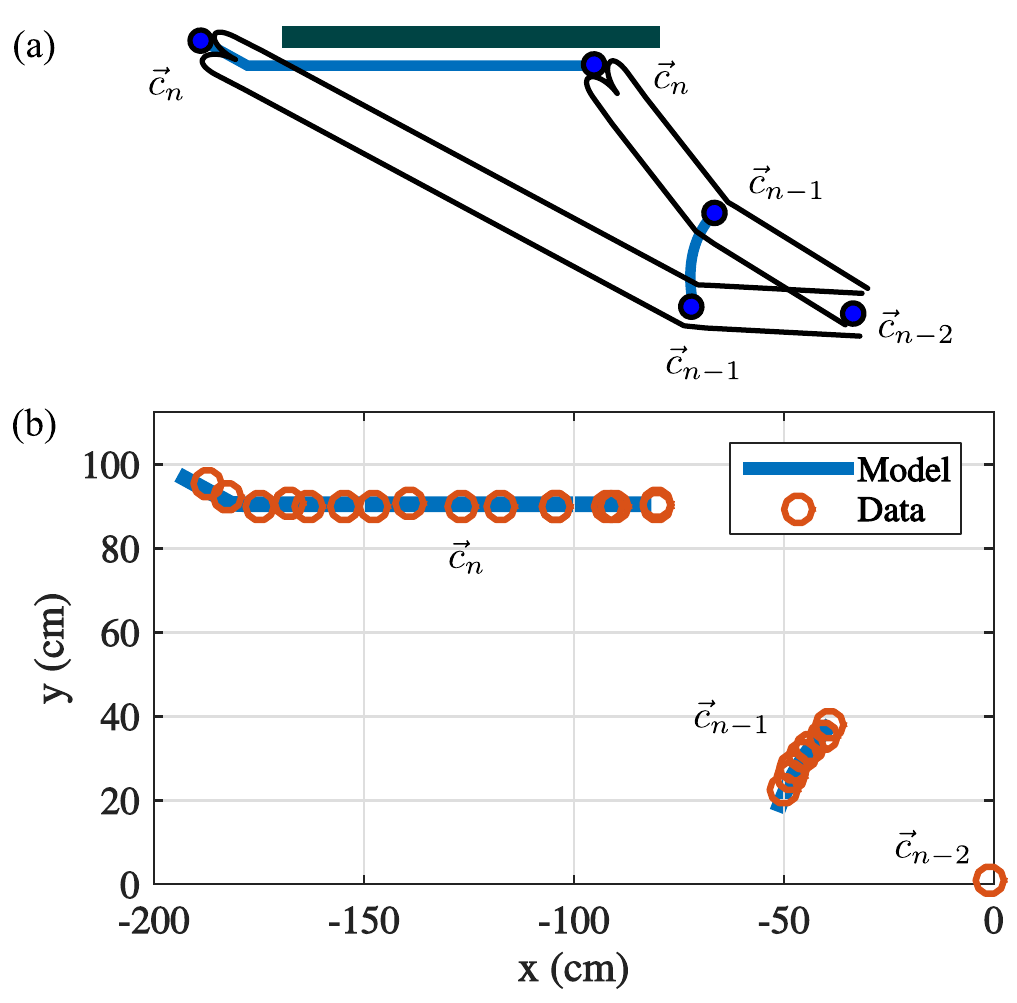}
\end{center}
\caption[Wall Pivot Point Trajectories]{Growth into a wall, pivoting about $\vec{c}_{n-2}$. Schematic of the interaction is shown in (a) and experimental data is compared to model prediction in (b). As expected, rather than pivoting about the most distal pivot point not in contact with the wall, $\vec{c}_{n-1}$, it pivots about $\vec{c}_{n-2}$ since $\vec{c}_{n-1}$ is a right-handed pivot point and the obstacle is pivoting the robot to the left.}
\label{fig:TrajectoriesWallPivotPoints} 
\vspace{-5mm}
\end{figure}

In the second experiment, a designed turn  to the right causes a robot to grow into a wall.  Due to the wall's angle relative to the most distal segment of the robot, the wall causes the robot to pivot to the left \mbox{(Fig.\ \ref{fig:TrajectoriesWallPivotPoints}(a))}. Because the most distal pivot point not in contact with the obstacle is right handed ($\vec{c}_{n-1}$) the obstacle pivot point is shifted one proximal to $\vec{c}_{n-2}$. As a result, both $\vec{c}_{n-1}$ and $\vec{c}_n$ rotate about $\vec{c}_{n-2}$ as the robot's tip slides along the wall. To test this second scenario, an overhead camera was used to track the three most distal pivot points $\vec{c}_n, \vec{c}_{n-1}$ and $\vec{c}_{n-2}$. Experimentally measured trajectories are compared against those predicted by the obstacle interaction model and shown in \mbox{Fig.\ \ref{fig:TrajectoriesWallPivotPoints}(b)}.

\begin{figure}
\begin{center}
\includegraphics[width=0.9\columnwidth]{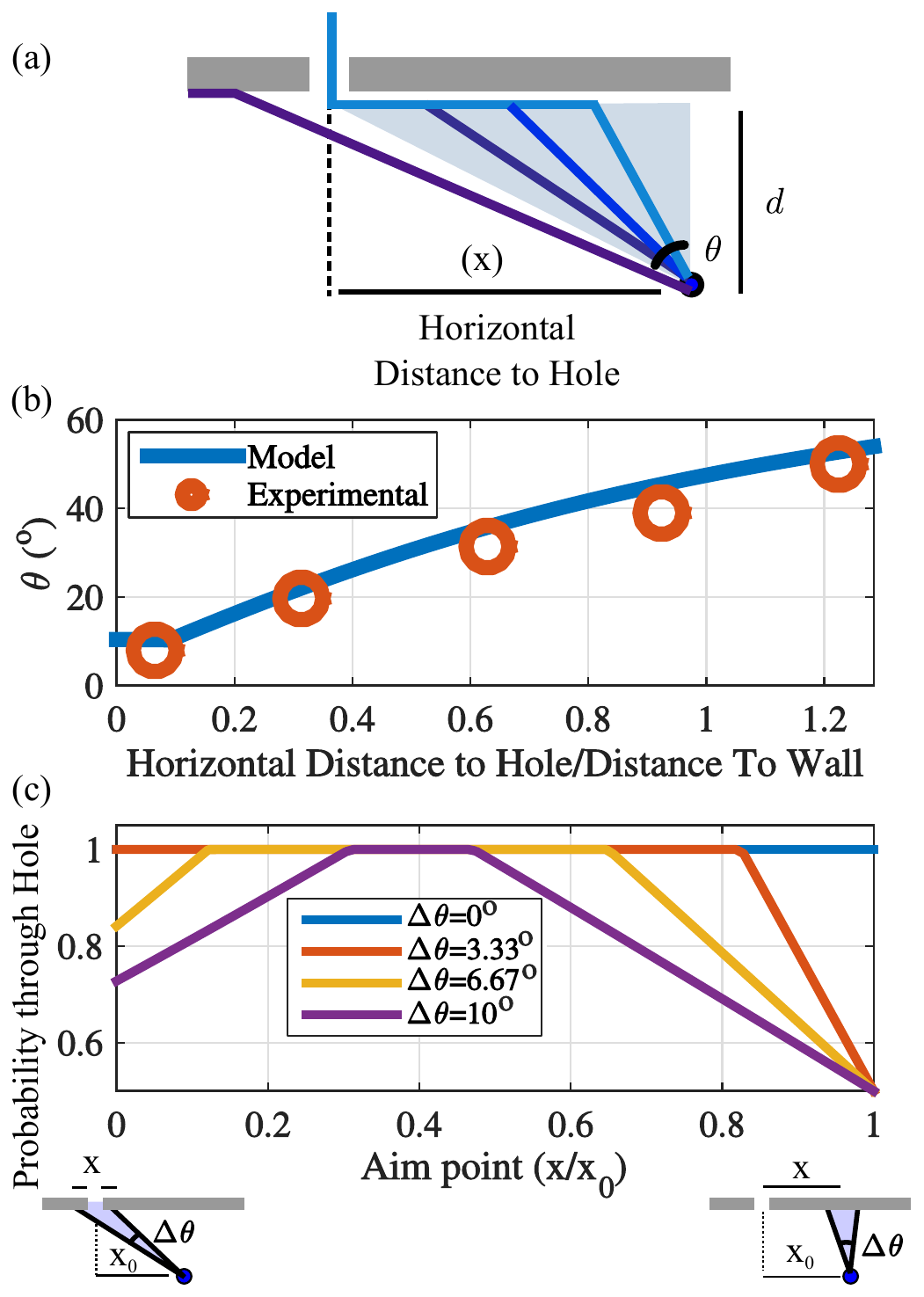}
\end{center}
\caption[Growth through Hole with Angular Uncertainty]{Experiment of growth through a hole-in-the-wall. (a) Several predicted trajectories of the tip of the soft growing robot for different approach angles are shown at a fixed distance from the hole. When the approach angle is within the light blue region with solid angle, $\theta$, the robot will grow through the hole. (b) Acceptable solid angle of initial orientations vs horizontal distance from hole. (c) Probability of successfully growing through a hole when there is uniform angular uncertainty ($\Delta \theta$) versus horizontal point the robot is nominally aimed at ($x$), from a fixed location $x_0$. This plot indicates that with any uncertainty, it is better to aim to the side of the hole than at it. Adapted from \citep{GreerICRA2018} \textcopyright\ IEEE 2018}
\label{fig:AngleRangeWall}
\vspace{-2mm}
\end{figure} 


\subsection{Growth Through a Hole in a Wall}
To understand how a planner can exploit environmental contact for navigation, we considered the task of navigating the robot through a small hole in a wall. This scenario is relevant in search and rescue, mining, and surgical applications. We studied this scenario by performing an experiment in which we repeatedly grew the robot through a hole in the wall, with a width of 6.5\,cm (\mbox{Fig.\ \ref{fig:AngleRangeWall}(a)}). From the Growth Along a Wall section, we know that if the robot is angled left of vertical, its tip will move along the wall to the left. Furthermore, the model predicts that if the ray extending from $\vec{c}_n$ in the direction of $\vec{c}_n-\vec{c}_{n-1}$ extends into the hole, it will grow through it. In this way, the obstacle serves to passively guide the robot's tip through the hole. Three predicted tip trajectories are shown in \mbox{Fig.\ \ref{fig:AngleRangeWall}(a)}.

For a fixed horizontal position, the model predicts that the robot will successfully grow through a hole if its starting orientation is within the shaded region in \mbox{Fig.\ \ref{fig:AngleRangeWall}(a)}. This region has starting orientations that range from just left of perpendicular to $\tan^{-1}(x/d)$ (aiming at the hole), where $x$ is the horizontal distance from the hole and $d$ is the vertical distance from the wall. \mbox{Fig.\ \ref{fig:AngleRangeWall}(b)} shows the range of starting orientations that will result in successfully growing through the wall-hole versus normalized horizontal distance from the wall for both the model and experimental trials. Fig.\ \ref{fig:AngleRangeWall}(c) suggests that in the case that there is uncertainty in the angle of approach, it is better to aim the robot at the wall closer to starting position than the location of the hole, rather than aiming directly at the hole.

\begin{figure}
\begin{center}
\includegraphics[width=0.8\columnwidth]{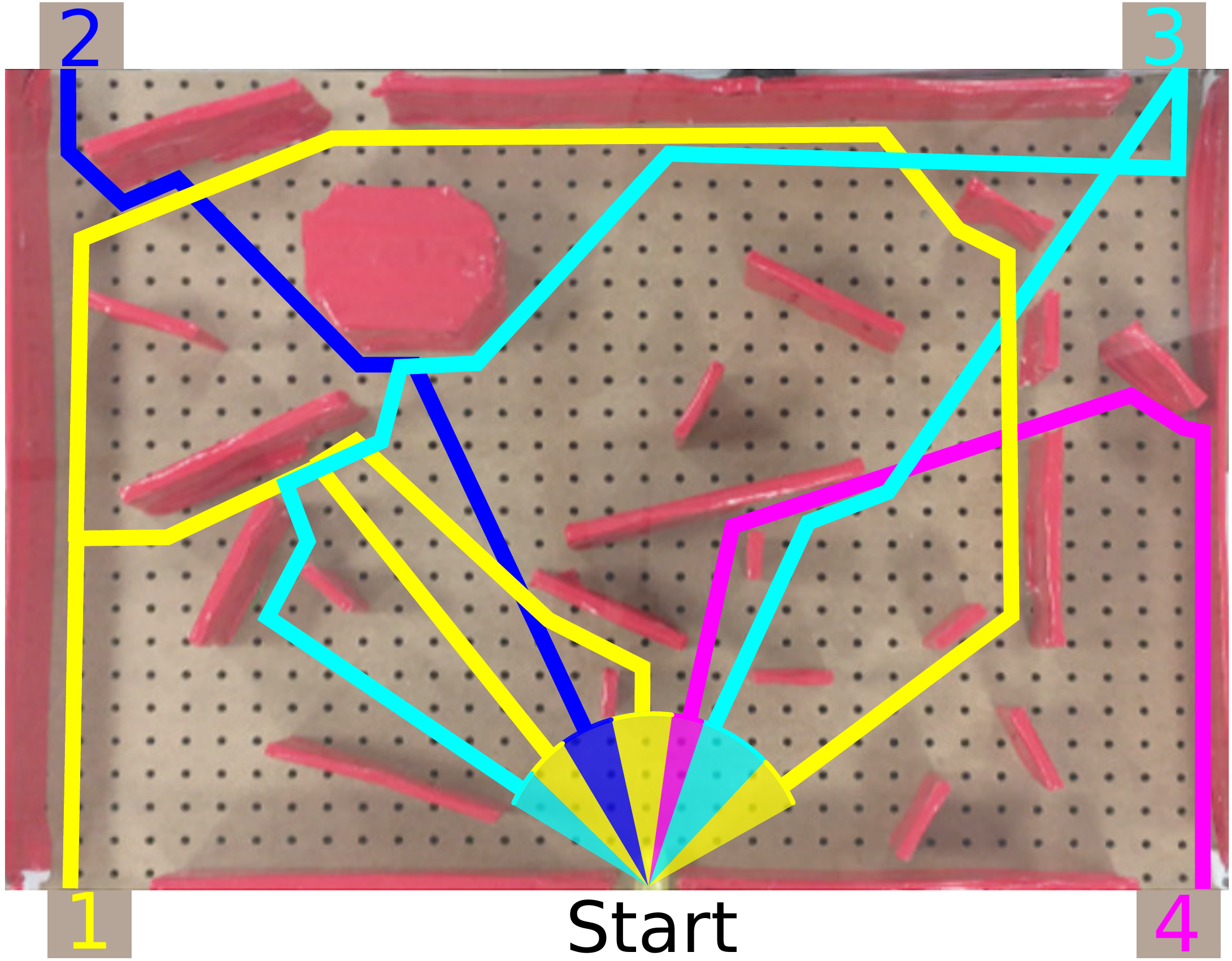}
\end{center}
\caption[Path Prediction Experimental Setup]{Obstacle course with tip trajectories computed using the obstacle interaction model. Four exit positions are labeled as 1, 2, 3, and 4 and correspond to colors yellow, blue, cyan, and pink, respectively. Depending on the starting orientation, the robot will end at one of the four locations. Varying over the possible starting orientations, there are seven transitions between ending points. Representative trajectories from each orientation regime is shown, colored to correspond to its ending location. Adapted from \citep{GreerICRA2018} \textcopyright\ IEEE 2018}
\label{fig:CoursePic}
\vspace{-5mm}
\end{figure}

\begin{figure*}[!t]
\begin{center}
\includegraphics[width=0.85\textwidth]{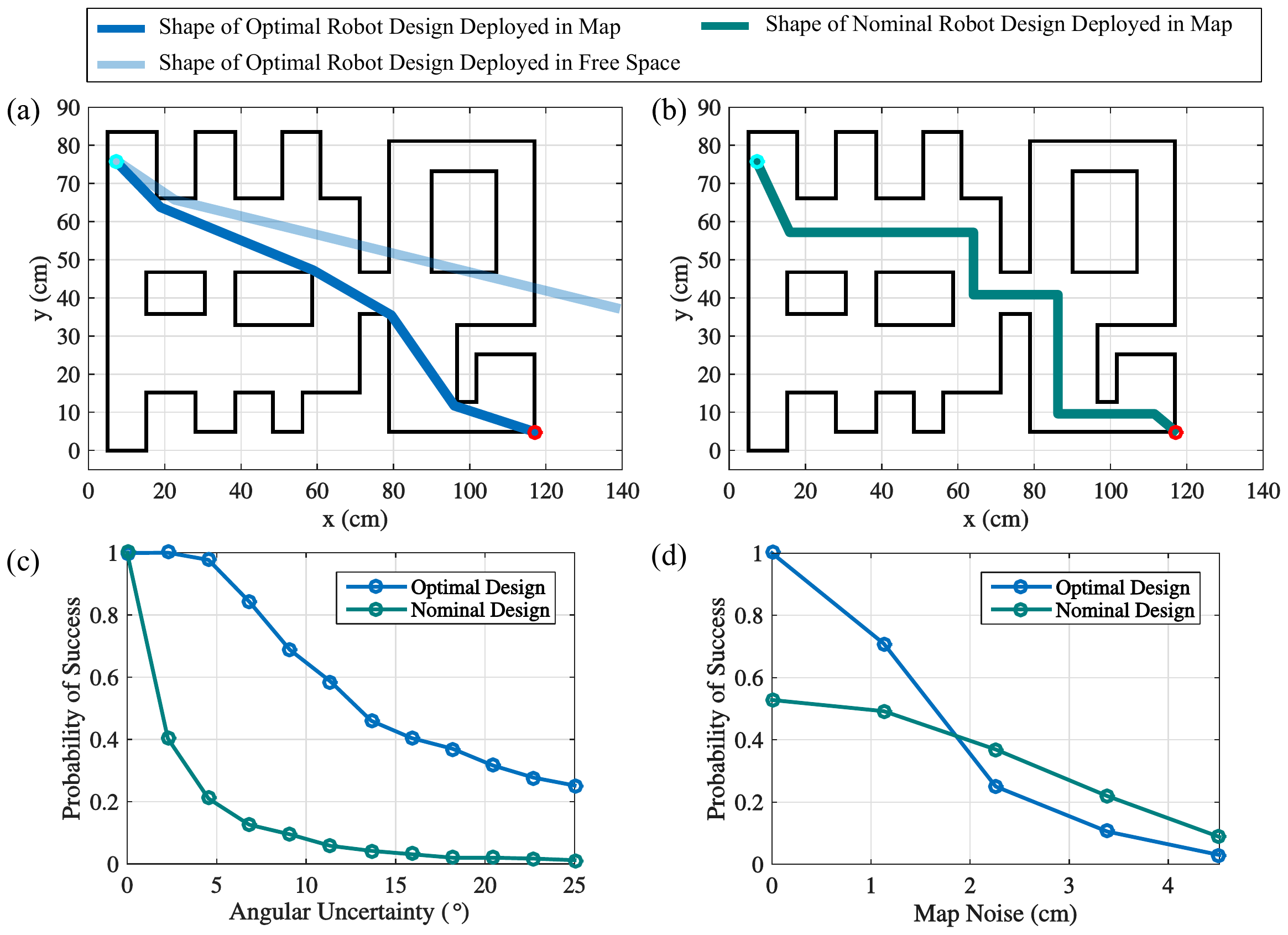}
\end{center}
\caption[Numerical Planning Results]{Path planning results. (a) Optimal design for navigating the tip of the soft growing robot from start location (top left) to destination location (bottom right). Free space deployment of the robot is shown in light blue and deployment of the robot in the map (calculated using the obstacle interaction model) is shown in dark blue. (b) Nominal design for the same task. This design maximizes distance from the robot's body to obstacles. (c) Probability of reaching the destination versus angular uncertainty for both the optimal design and nominal design. Utilizing obstacle interactions increases robustness of the robot design. (d) Probability of reaching destination with varying map uncertainty. Nominal design is less affected by imperfect map knowledge since design is chosen to maximize distance from its environment. }
\label{fig:PlanningResults}
\end{figure*}

\begin{figure}
\begin{center}
\includegraphics[width=1\columnwidth]{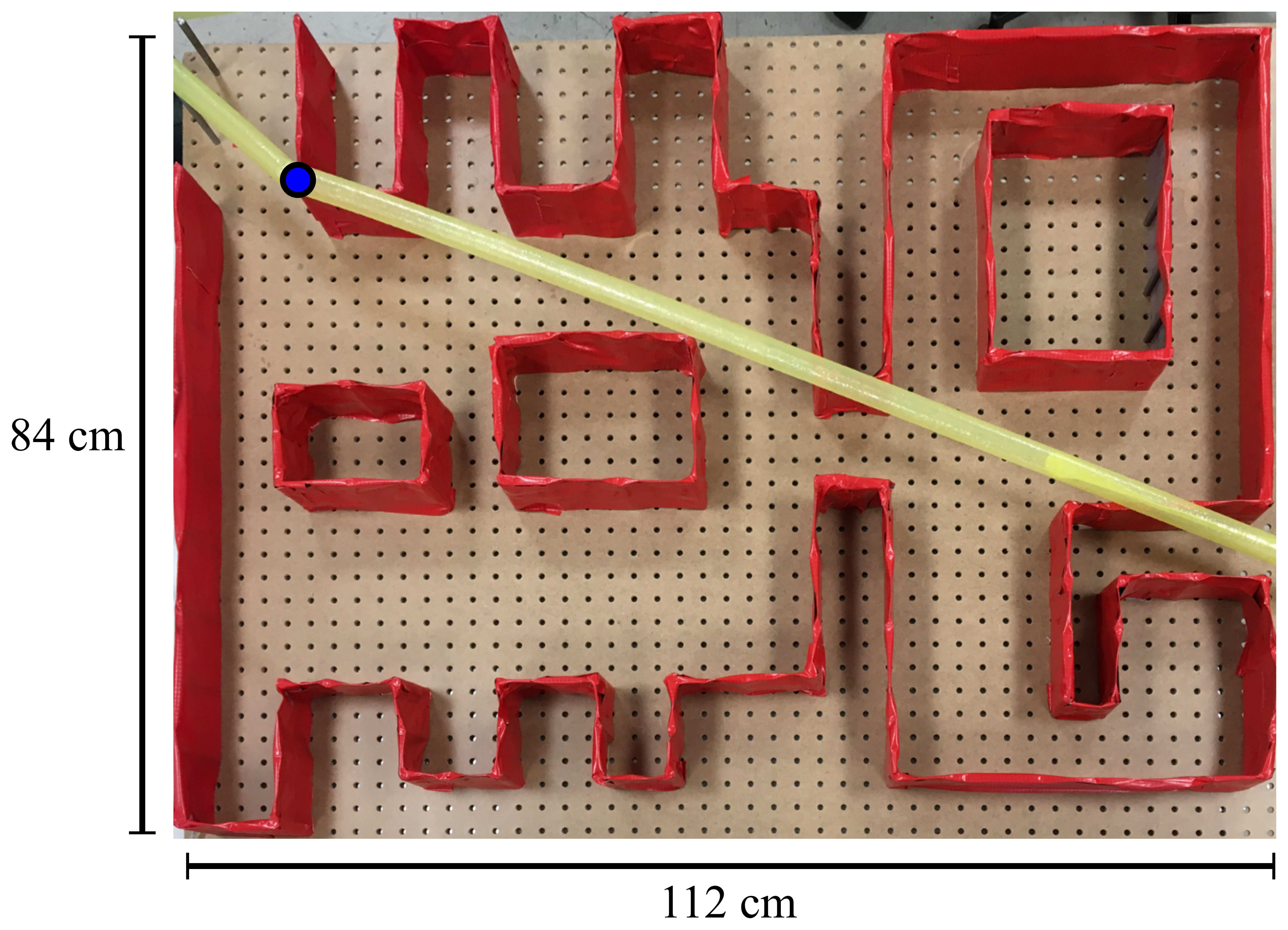}
\end{center}
\vspace{-5mm}
\caption[Free Space Deployment of Optimal Robot Design]{Free space deployment of optimal robot design shown in \mbox{Fig.\ \ref{fig:PlanningResults}(a)}. Robot is above the plane of the map in this image. The design consists of one $18^\circ$ left turn at a length of 18.8 cm; the designed turn marked with a blue circle. The physical implementation of the turn was more shallow than the design prescribed. Nonetheless, the deployed robot reached the desired destination as shown in the second row of \mbox{Fig.\ \ref{fig:PlanningSequence}.}}
\label{fig:FreeSpace}
\end{figure}

\begin{figure*}
\begin{center}
\includegraphics[width=\textwidth]{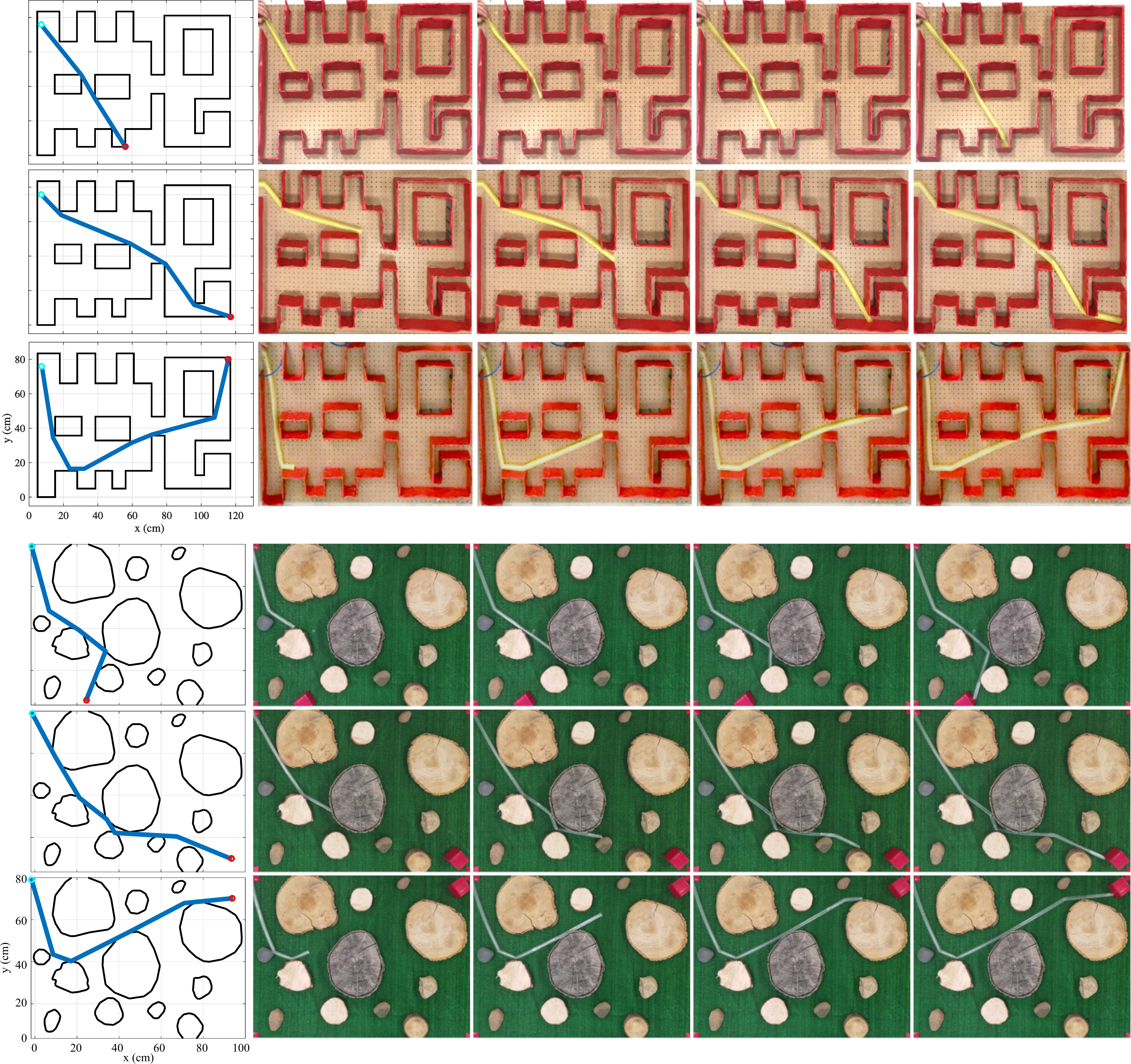}
\end{center}
\caption[Physical Test of Optimal Design]{Physical implementation of optimal robot designs. Each row corresponds to a different destination and for each destination, an optimal design was produced by the planning method presented in this paper. The left column contains the path predicted by the obstacle interaction model and the columns to the right show four snapshots during growth to the destination. The second row is an implementation of the optimal robot design shown in \mbox{Fig. \ref{fig:PlanningResults}(a)}.}
\label{fig:PlanningSequence} 
\end{figure*}

\subsection{Growth Through a Cluttered Environment}
To demonstrate the obstacle interaction model, we created a planar environment with obstacles to grow the robot through and compared the observed robot trajectory with the result predicted by the path computation algorithm. The robot was made out of 2.54 cm wide and 51 $\mu$m thick thin-walled polyethylene tubing and driven using compressed air that ranged in pressure from 7 and 21 kPa.  \mbox{Fig.\ \ref{fig:CoursePic}} shows the obstacle course that was used, which was 122 cm wide by 92 cm high. It has four possible exits that are labeled 1, 2, 3, and 4. 

By starting at the same position, but varying the orientation of the soft growing robot, its path is changed. Sweeping the starting orientation over the range of possible angles, the ending location changes six times. For example, moving the starting orientation from vertical to just right of vertical changes the ending location from the lower-left corner (1) to the lower-right corner (4). Representative tip trajectories predicted by the obstacle interaction model are overlaid in \mbox{Fig.\ \ref{fig:CoursePic}}. The path computation algorithm correctly predicted the exit location of seven out of seven trials.

\subsection{Optimal Robot Design}
To evaluate the performance of the path planning method, we adapted a map from the grid-based pathfinding dataset \citep{sturtevant2012benchmarks} for use with our path planner. The adapted map is shown in \mbox{Fig.\ \ref{fig:PlanningResults}(a) and (b)}. The starting and destination locations for the robot are in the upper left and lower right corners of the map, respectively and indicated by teal (starting location) and red (destination location) circles. 

First, numerical experiments were performed to test the efficacy of the proposed planning method. \mbox{Fig.\ \ref{fig:PlanningResults}(a)} shows the nominal robot design that was output by the method in both a free space (light blue) and deployed (dark blue) configuration. As can be seen, the free space configuration differs from the deployed configuration because obstacles are used to redirect the robot to its destination. The generated robot design was contrasted with a robot design that avoids obstacles \mbox{(Fig.\ \ref{fig:PlanningResults}(b))}. This design was chosen to approximately maintain as far a distance from the obstacles as possible while reaching the destination. Because the nominal design does not interact with the environment, the free space and deployed robot configuration for this design are identical. 

\mbox{Fig.\ \ref{fig:PlanningResults}(c)} compares the robustness of the two designs to uncertainty in their physical realizations. The data was generated by running 10000 Monte Carlo simulations for varying amounts of uncertainty in the robot design. A simulation was counted as a success if the robot tip ended within 5 cm of the destination and a failure otherwise. The numerical simulations indicate that the proposed planning method is significantly more reliable for all amounts of uncertainty. \mbox{Fig.\ \ref{fig:PlanningResults}(d)} compares the effect map uncertainty has on the two designs. A separate set of 10000 Monte Carlo simulations were run on maps that were perturbed from the nominal map by adding normally distributed random noise with varying standard deviation (x-axis) to each node of the nominal map. A fixed amount of design uncertainty was used for all trials ($\sigma_\theta=2^\circ$ and $\sigma_l = $1.1cm). Unsurprisingly, the nominal design was less affected by map uncertainty than the optimal design since the nominal design was made to avoid its environment, where as the optimal design was made to use its environment.

In addition to numerical tests, we performed physical experiments in two types of environments: a maze-like environment and a forest-like environment. The first environment is a physical realization of the map in \mbox{Fig. \ref{fig:PlanningResults}} and is shown in \mbox{Fig. \ref{fig:FreeSpace}} along with an implementation of the design output by the planning algorithm in \mbox{Fig. \ref{fig:PlanningResults}(a)}. Turns were made by taping a pinch into the side of the body of the robot as shown in \mbox{Fig. \ref{fig:RobotOverview}}.  

\mbox{Fig.\ \ref{fig:PlanningSequence}} shows deployments of the robot to different goal points within the two environments. In particular, the second row contains stills taken from a successful deployment of the robot shown in \mbox{Figs.  \ref{fig:PlanningResults} and \ref{fig:FreeSpace}}. The robot makes contact with a wall, which was not predicted in the simulation of the nominal design (\mbox{Fig.\ \ref{fig:PlanningResults}(a)}). This was due to mismatch between the physical and virtual maps. Despite both the robot manufacturing error and mismatch between the physical and virtual maps, the robot successfully reached the destination. The bottom three rows of \mbox{Fig. \ref{fig:PlanningSequence}} show deployments of the robot in the second, forest-like, environment. It contains irregularly shaped obstacles and tests the model and planning method's applicability to natural environments with more complicated robot-obstacle interactions. As shown in \mbox{Fig. \ref{fig:Main}}, similar benefits are found by exploiting obstacle contact, namely passive turning by the environment and reduction in uncertainty of the robot's motion.

\section{Conclusion}
\label{sec:Conclusion}
For a robot moving through a cluttered environment, it is inevitable that the robot will interact with obstacles. 
Rather than seeing these obstacle interactions as inherently negative, we show that they can be advantageous for navigating the soft growing robot to a particular destination because interactions with obstacles can consolidate many possible paths down to a single desired path, and these interactions can direct the robot to locations not on a straight line path from its starting point. This work describes both a model and a planning method to exploit robot-obstacle interactions for navigation of a soft growing robot. Though this principle was shown for the specific case of a soft growing robot, it applies more broadly to any robot that passively follows the contour of an obstacle. 



Our obstacle interaction model has several limitations. An assumption of head-on contact (Planar Kinematic Model section) is that the robot's backbone will pivot about a pivot point as its tip slides along an obstacle contour. This is only true when the robot (i) buckles and (ii) the cause of buckling is a transverse rather than axial load. These assumptions are satisfied when the membrane material is sufficiently thin, air pressure in the backbone is low, the free length is short, and the angle of contact is above a few degrees. If these are not true, the robot will either bend, or buckle at a point other than the predicted pivot point. Though it will not affect the accuracy of the predicted tip location for a single obstacle, it could affect the tip predictions for multiple, chained obstacle interactions.

\section*{Acknowledgment}

This work was supported in part by the National Science Foundation (grant no. 1637446) and Air Force Office of Scientific Research (grant no. FA2386-17-1-4658).

\bibliographystyle{SageH}
\bibliography{ObstaclePlanning}

\end{document}